\documentclass[lettersize,journal]{IEEEtran}
\usepackage{color,soul} 
\soulregister\cite7 % 针对\cite命令 
\soulregister\citep7 % 针对\citep命令 
\soulregister\citet7 % 针对\citet命令 
\soulregister\ref7 % 针对\ref命令 
\soulregister\pageref7 % 针对\pageref命令
\usepackage{amsmath,amsfonts}
\usepackage{array}
\usepackage{booktabs}
\usepackage[caption=false,font=normalsize,labelfont=sf,textfont=sf]{subfig}
\usepackage{textcomp}
\usepackage{stfloats}
\usepackage{verbatim}
\usepackage{graphicx}
\usepackage{cite}
\usepackage{multirow}
\usepackage[pagewise]{lineno}
\usepackage[table,xcdraw]{xcolor}
\usepackage[numbers]{natbib}
\usepackage{float}

\usepackage{amsthm}
\usepackage{amsmath}
\usepackage{multirow}
\usepackage{soul}
\usepackage{svg}
\usepackage{algorithmic}
\usepackage{makecell}

\usepackage{threeparttable}
\usepackage[linesnumbered,ruled]{algorithm2e}
\usepackage{tikz}

\usepackage{amssymb}
\usepackage{bm}
\usepackage{balance}
\usepackage{url}
\usepackage[colorlinks=true,
            citecolor=blue,
            linkcolor=blue,
            anchorcolor=blue,
            urlcolor=blue]{hyperref}

\newtheorem{thm}{Theorem}
\newtheorem{rmk}{Remark}

\usepackage[linesnumbered,ruled]{algorithm2e}

\begin{document}

%\multirowsetup{vcenter}
% \pagewiselinenumbers
% \switchlinenumbers

\title{Graph Optimality-Aware Stochastic LiDAR Bundle Adjustment with Progressive Spatial Smoothing}

\author{
\IEEEauthorblockN{
Jianping Li, \textit{Member}, \textit{IEEE},
Thien-Minh Nguyen,
Muqing Cao,
Shenghai Yuan, \textit{Member}, \textit{IEEE},
Tzu-Yi Hung,
Lihua Xie, \textit{Fellow}, \textit{IEEE}
}
\vspace{-20pt}
\thanks{
J. Li, T. Nguyen, S. Yuan, L. Xie are with School of Electrical and Electronic Engineering, Nanyang Technological University, Singapore 639798. M. Cao is with Carnegie Mellon University, Pittsburgh, PA 15213. T. Hung is with Delta Research Center, Delta Electronics, Singapore, 339274.
This research was conducted under project WP5 within the Delta-NTU Corporate Lab with funding support from A*STAR under its IAF-ICP program (Grant no: I2201E0013) and Delta Electronics Inc.
}

}

% The paper headers
\markboth{Journal of \LaTeX\ Class Files,~Vol.~xxx, No.~xxx, xxx~xxx}%
{Shell \MakeLowercase{\textit{et al.}}:Graph Optimality-Aware Stochastic LiDAR Bundle Adjustment with Progressive Spatial Smoothing}

\maketitle

\begin{abstract}

Large-scale LiDAR Bundle Adjustment (LBA) to refine sensor orientation and point cloud accuracy simultaneously to build the navigation map is a fundamental task in logistics and robotics. Unlike pose-graph-based methods that rely solely on pairwise relationships between LiDAR frames, LBA leverages raw LiDAR correspondences to achieve more precise results, especially when initial pose estimates are unreliable for low-cost sensors. However, existing LBA methods face challenges such as simplistic planar correspondences, extensive observations, and dense normal matrices in the least-squares problem, which limit robustness, efficiency, and scalability. To address these issues, we propose a Graph Optimality-aware Stochastic Optimization scheme with Progressive Spatial Smoothing, namely PSS-GOSO, to achieve \textit{robust}, \textit{efficient}, and \textit{scalable} LBA. The Progressive Spatial Smoothing (PSS) module extracts \textit{robust} LiDAR feature association exploiting the prior structure information obtained by the polynomial smooth kernel. The Graph Optimality-aware Stochastic Optimization (GOSO) module first sparsifies the graph according to optimality for an \textit{efficient} optimization. GOSO then utilizes stochastic clustering and graph marginalization to solve the large-scale state estimation problem for a \textit{scalable} LBA. We validate PSS-GOSO across diverse scenes captured by various platforms, demonstrating its superior performance compared to existing methods. Moreover, the resulting point cloud maps are used for automatic last-mile delivery in large-scale complex scenes. The project page can be found at: \url{https://kafeiyin00.github.io/PSS-GOSO/}.

\end{abstract}


\section{Introduction}

LiDAR Bundle Adjustment (LBA) is a fundamental task in logistic \citep{guan2015using} and robotics \citep{liu2023large}, aiming to simultaneously refine sensor orientation and point cloud map accuracy. As a critical step for accurate 3D mapping of the complex scenes, LBA reduces point cloud discrepancies and enhances consistency, which is essential for downstream applications such as infrastructure maintenance \citep{qiu2024whu, lin2022semantic}, and robotic last-mile delivery utilizing prior maps \citep{gan2023dp,li2024hcto}.

\begin{figure*}[]
  \centering
  \includegraphics[width=\linewidth]{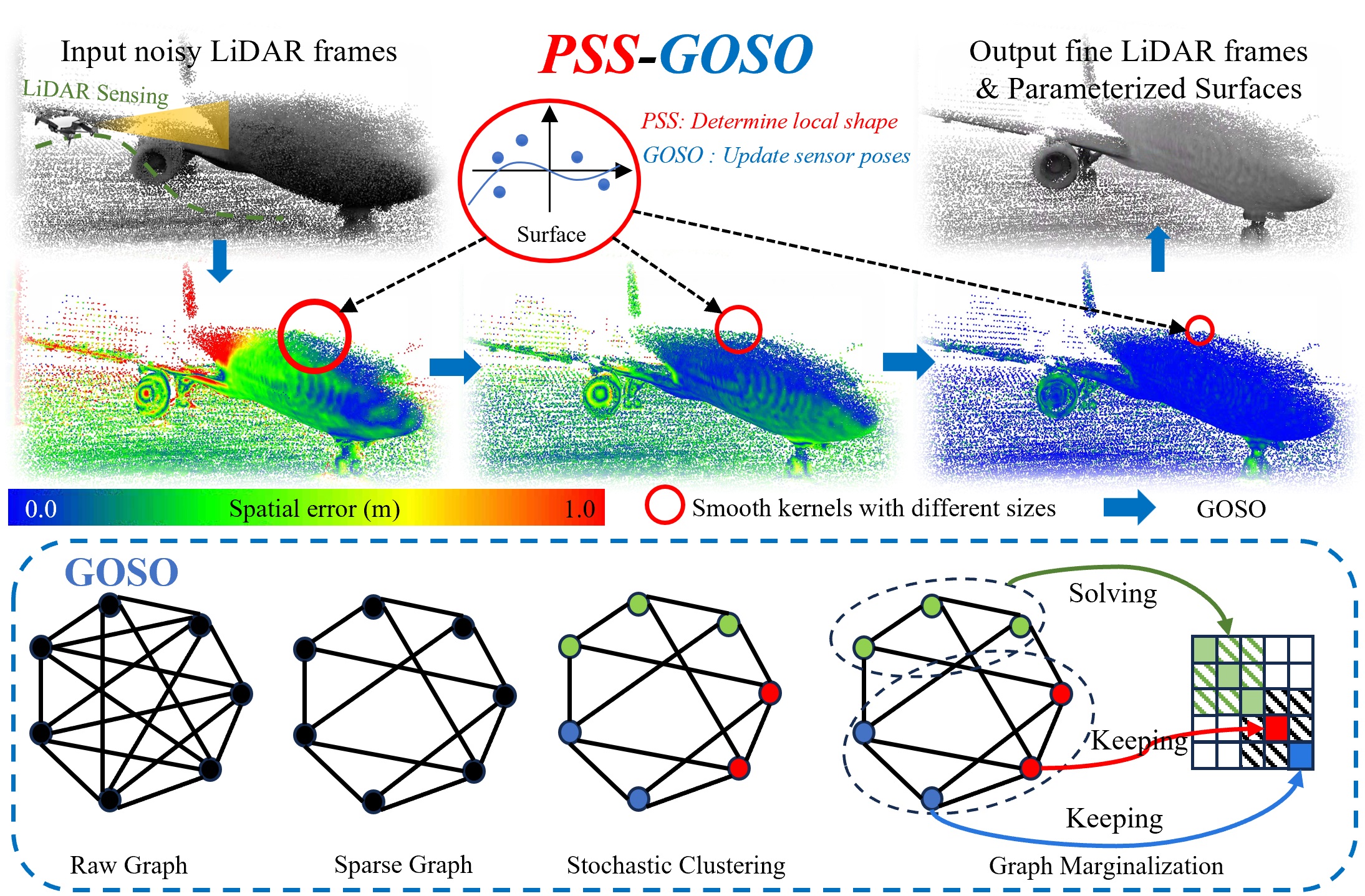}
  \caption{Graph optimality-aware stochastic optimization with progressive spatial smoothing for a \textit{robust}, \textit{efficient}, and \textit{scalable} LBA. The Progressive Spatial Smoothing (PSS) module extracts \textit{robust} LiDAR feature association exploiting the prior structure information obtained by the polynomial smooth kernel. The Graph Optimality-aware Stochastic Optimization (GOSO) module first satisfies the graph according to optimality for an \textit{efficient} optimization. GOSO then utilizes stochastic clustering and graph marginalization to solve the large-scale state estimation problem for a \textit{scalable} LBA.}
  \label{fig:abstract}
  \vspace{-15pt}
\end{figure*}

LBA has been extensively studied over the past decades and has been applied in traditional mobile mapping systems, such as aerial laser scanning \citep{glira2015rigorous} and vehicle-borne laser scanning \citep{li2023mucograph}. However, with the rise of compact, low-cost 3D sensing systems and the increasing demand for 3D mapping in complex environments denied by GNSS, existing LBA methods are not directly applicable to these low-cost platforms due to issues related to robustness, efficiency, and computational load. Consequently, adapting LBA for compact sensing systems in large-scale, complex scenes has become a prominent topic of interest in both academic and industrial communities.

% related works
Existing solutions for refining LiDAR orientation can be categorized into two main classes based on their optimization strategies: indirect optimization-based methods and direct optimization-based methods. A brief review of these categories is provided below.

\subsection{Indirect optimization-based method} 

Indirect optimization-based methods first extract relative transformations by performing pairwise registration techniques, such as ICP \citep{segal2009generalized}, NDT \citep{einhorn2015generic}, and their variants \citep{koide2021voxelized}, and then optimize poses without involving the raw correspondences used in pairwise registration. For example, \cite{dong2018hierarchical} registers unordered terrestrial laser scanning point clouds by analyzing the pair-wise relationships between laser frames, integrating the point clouds incrementally without redundant pose constraints. This approach can be seen as one of the simplest indirect optimization-based methods to merge multiple point clouds. However, the incremental fusion approach may lead to error accumulation, which is often mitigated by using pose graph optimization \citep{carlone2014fast}. A critical challenge in accurate pose graph optimization is transforming pairwise registration results into relative pose constraints that follow a Gaussian distribution \citep{yang2022hierarchical}. In practice, this Gaussian assumption is often overestimated \citep{koide2022globally}, which can result in information loss and inaccuracies. Additionally, pose graph optimization may produce erroneous results if the relative pose estimation is unreliable in areas with low frame overlap. Indirect optimization-based methods are not the primary focus of this paper.

\subsection{Direct optimization-based method} 
% LBA 的方式
Direct optimization-based methods extract feature correspondences between LiDAR frames and simultaneously optimize both poses and point clouds. Compared to indirect optimization-based methods, direct optimization often achieves higher accuracy by avoiding information loss during the relative transformation estimation step inherent in indirect methods. However, three main challenges need to be addressed in direct optimization-based methods: \textbf{feature association}, \textbf{efficiency}, and \textbf{scalability}.

% 特征
\textbf{Feature association} in LBA determines the type of correspondence used for pose correction. Commonly utilized features include points \citep{li2023mucograph}, planes \citep{skaloud2006rigorous}, and their combinations \citep{liu2021balm}. While these features perform effectively in most structured environments, their performance can degrade and lead to divergence in complex environments that lack clear structural features. Despite the effectiveness of these commonly used features, existing work, PSS-BA \citep{li2024pss}, demonstrated that spatial smoothing can enhance feature association by providing more robust and richer constraints in complex scenarios. Nevertheless, PSS-BA \citep{li2024pss} is constrained by data volume limitations and is not well-suited for large-scale environments. Consequently, this work aims to extend PSS-BA to enhance its efficiency and scalability.

% 效率
\textbf{Efficiency} is crucial for end users, particularly in large-scale mapping tasks. Many existing pose optimization techniques face a trade-off between accuracy and efficiency \citep{li2024hcto}. For example, \cite{jiang2017efficient} accelerates the adjustment of the image bundle by taking advantage of the graph structure of poses and employing the maximal expansion of the spanning tree. Similarly, \cite{li2024hcto} enhances real-time performance in continuous state estimation by optimizing the normal matrix. However, to the best of our knowledge, no current LBA methods effectively select the most useful constraints from redundant correspondences to improve the efficiency of the optimization process while maintaining comparable accuracy.

% 参数，大范围
\textbf{Scalability} is a critical issue for LiDAR Bundle Adjustment (LBA), especially in large-scale mapping applications. Traditional mobile mapping systems equipped with high-accuracy Positioning and Orientation Systems (POS) that work with GNSS signals can model the small orientation errors from the high-accuracy POS as splines for correction \citep{poppl2024flexible}. In such cases, a relatively small number of feature correspondences can suffice for achieving high-quality results in large-scale applications. The sparse normal matrix used in the least-square estimation resulting from a small number of features is manageable with sparse-solving tools. However, with the rise of low-cost sensors used for 3D mapping in complex environments denied GNSS, scalability becomes more challenging. Low-cost LiDAR SLAM often provides less accurate initial poses, whose errors cannot be effectively modeled using splines to reduce system states. Additionally, the dense correspondences used for optimization result in a very dense normal matrix, which limits the scalability of existing LBA methods. To address these issues, \cite{yang2022hierarchical} introduced a scan-block structure to segment LiDAR frames from UAV systems, improving data quality. Similarly, \cite{liu2023large} proposed a hierarchical adjustment scheme that splits LiDAR frames into blocks based on time, thus mitigating data volume limitations. However, these sequential splitting strategies \citep{liu2023large} do not fully account for the spatial relationships in the pose graph and suffer from non-rigid drift in local blocks, leading to discrepancies in the final point clouds \citep{li2023whu}. Moreover, such sequential splitting methods are not well-suited for multi-session tasks, such as integrating scanning data collected from different platforms for large-scale scenes \citep{nair2024hilti}.

To achieve a \textit{robust}, \textit{efficient}, and \textit{scalable} LiDAR Bundle Adjustment (LBA), the core concept of the proposed PSS-GOSO is illustrated in Fig. \ref{fig:abstract}. The main contributions of PSS-GOSO can be summarized as follows:
\begin{itemize}
    \item PSS-GOSO enhances LiDAR feature association by leveraging prior structural information obtained through a polynomial smooth kernel, resulting in improved accuracy compared to commonly used planar feature associations.
    \item PSS-GOSO optimizes efficiency by sparsifying the relation graph based on optimality, while still maintaining comparable accuracy.
    \item PSS-GOSO employs optimality-aware stochastic optimization and graph marginalization techniques to address large-scale state estimation challenges, effectively considering spatial relationships and achieving scalable LBA.
    \item The validation and effectiveness of PSS-GOSO are demonstrated through extensive experiments on both public and proprietary large-scale cross-platform datasets, including UGVs, UAVs, and wearable devices.
\end{itemize}

The remainder of this paper is organized as follows: Section \ref{section_pre} provides the preliminary details for our system. The Progressive Spatial Smoothing (PSS) module, which is used for robust LiDAR feature association, is described in Section \ref{section:pssba}. Section \ref{section:goso} presents a detailed description of the Graph Optimality-aware Stochastic Optimization (GOSO) method. Section \ref{section:experiments} covers the experiments conducted using both public and proprietary datasets. Finally, Section \ref{section:conclusion} offers conclusions and discusses future work.
\section{Preliminary \label{section_pre}}
\subsection{Notation}

\subsubsection{Poses} In this paper, we use italic, bold lowercase, and bold uppercase letters to represent scalars, vectors, and matrices, respectively. Three main frames of reference are used in our proposed method, namely, the world frame $F^{W}$, the LiDAR frame $F^{L}$, and the tangent frame $F^{S}$ defined in a local 3D space \citep{dong2017novel}. We denote any point observed by the LiDAR in the $i$-th frame as $\mathbf{p}^{L}_i$. The pose for the $i$-th LiDAR frame is denoted as $\mathbf{X}_i=[\mathbf{R}_i,\mathbf{t}_i]$, where $\mathbf{R}_i \in \mathrm{SO(3)}$ is the rotation matrix, $\mathbf{t}_i \in \mathbb{R}^3$ is the translation vector. All the poses are denoted as $\mathbf{X}=\{\mathbf{X}_i\}^m_{i=1}$.
The initial value is noted with breve $\breve{\circ}$, and the estimated value is noted with the hat $\hat{\circ}$.
The estimated pose of the $i$-th LiDAR frame is denoted as $\hat{\mathbf{X}}_i=[\hat{\mathbf{R}}_i=\mathrm{Exp}{(\Delta \bm{\theta}_i)}\mathbf{R},\hat{\mathbf{t}}_i=\Delta \mathbf{t}_i+\mathbf{t}_i]$, 
where $\Delta\bm{\theta}_i \in \mathbb{R}^3 $ and $\Delta \mathbf{t}_i \in \mathbb{R}^3$ are the corresponding errors. The exponential map $\mathrm{Exp}:\mathbb{R}^3 \rightarrow \mathrm{SO(3)}$ has the form:
\begin{equation}
    \mathrm{Exp}(\Delta\bm{\theta}_i) = \mathbf{I} + \frac{\mathrm{sin}(\Vert\Delta\bm{\theta}_i\Vert)}{\Vert\Delta\bm{\theta}_i\Vert}[\Delta\bm{\theta}_i]_{\times} + \frac{1-\mathrm{cos}(\Vert\Delta\bm{\theta}_i\Vert)}{\Vert\Delta\bm{\theta}_i\Vert^{2}}[\Delta\bm{\theta}_i]^{2}_{\times}. \label{eq:Exp}
\end{equation}
The point $\hat{\mathbf{p}}_{i}^{W}$ in the world frame $F^{W}$ can be obtained by:

\begin{equation}
    \hat{\mathbf{p}}_{i}^{W}=\hat{\mathbf{R}}_i\mathbf{p}_{i}^{L}+\hat{\mathbf{t}}_i, \label{eq:projection}
\end{equation}

\subsubsection{3D Normal} For an initial imperfect normal $\hat{\mathbf{n}}_i$ associated with $\hat{\mathbf{p}}_{i}^{W}$, it satisfies $ \Vert\hat{\mathbf{n}}\Vert=1 $. Thus, the normal vector's degree of freedom is two, and it can be written as:
\begin{equation}
    \hat{\mathbf{n}}_i \approx \mathbf{n}_i + [\hat{\mathbf{n}}^0_i,\hat{\mathbf{n}}^1_i]\Delta \bm{\phi}_i, \label{eq:normal_error}
\end{equation}
where $\Delta \bm{\phi}_i$ with dimension of $2 \times 1$ represents the small errors. The vectors $\hat{\mathbf{n}}^0_i,\hat{\mathbf{n}}^1_i$ are two unit vectors that are orthogonal to both $\hat{\mathbf{n}}_i$ and each other.

\subsubsection{Polynomial Smoothing Kernel} For a second-order polynomial surface defined within a local tangent space $F^{S}$, its functional form is as follows:

\begin{equation}
    z = f\left( x, y \right) = \bm{\alpha}^{\top}\left[ x^2, y^2,xy,x,y \right] ^{\top}, \label{eq:polynomial}
\end{equation}
where the vector $\bm{\alpha}$ with dimension of $5 \times 1$ represents the coefficients that describe the surface's shape. Assuming a continuous 3D environment, the point clouds within the local tangent space are projected onto the surface $f$ to mitigate measurement noise.

\subsection{LiDAR Bundle Adjustment}

A LBA problem can be built upon a relation graph $\mathbf{G}=(\mathbf{X},\mathbf{E})$. Here, $\mathbf{X}$ is the set of $m$ LiDAR poses. $\mathbf{E}=\{\mathbf{E}_k\}_{k=1}^l$ denotes the $l$ correspondences between the LiDAR poses. The feature correspondences can be the point-to-point, point-to-line, and point-to-planarity styles \citep{zhang2014loam, segal2009generalized}. In this work, we use the polynomial smoothing kernel \eqref{eq:polynomial} to formulate the correspondences, which will be detailed in Section \ref{section:pssba}. Generally, the feature correspondence $\mathbf{E}_k$ depends on the actual poses $\mathbf{X}$, i.e. $\mathbf{E}_k = \mathrm{E}(\mathbf{X})_k$. In essence, LBA is the following minimization problem \eqref{eq:bundle_adjustment}.
\begin{equation}
  \underset{\mathbf{X}}{\mathrm{argmin}}~\mathrm{F}(\mathbf{X})= \Vert \mathrm{E}(\mathbf{X})\Vert^2, \label{eq:bundle_adjustment}
\end{equation}
where $\mathrm{E}(\mathbf{X})$ is the concatnation of $\{\mathrm{E}(\mathbf{X})_k\}^{l}_{k=1}$. Setting the value of $\mathbf{X}$ with the initial guess $\mathbf{\mathbf{\breve X}}$, the Levenberg Marquardt (LM) algorithm \citep{ranganathan2004levenberg} achieves a pose correction step $\Delta \mathbf{X}$ at each iteration by linearizing $\mathrm{F}(\mathbf{\mathbf{\breve X}})$ as \eqref{eq:linearing}. 
\begin{equation}
  \underset{\Delta \mathbf{X}}{\mathrm{argmin}}~\mathrm{F}(\mathbf{X}) = \Vert\mathbf{J} \Delta \mathbf{X} + \mathrm{E}(\mathbf{\breve X}) \Vert^2+\lambda \Vert \mathbf{D} \Delta \mathbf{X} \Vert^2, \label{eq:linearing}
\end{equation}
where $\mathbf{J}$ is the Jacobian matrix. The solution of \eqref{eq:linearing} is as follows:
\begin{equation}
  \underset{\mathbf{H}}{\underbrace{(\mathbf{J}^{\top}\mathbf{J}+\lambda \mathbf{D}^{\top}\mathbf{D})}}\Delta \mathbf{X}=\underset{\mathbf{y}}{\underbrace{-\mathbf{J}^{\top}\mathrm{E(}\breve{\mathbf{X}})}} \label{eq:lm},
\end{equation}
where $\lambda$ is the damping factor, and $\mathbf{D}=\mathrm{diag}(\mathbf{J}^{\top}\mathbf{J})^{1/2}$.

\subsection{Optimality of LiDAR Bundle Adjustment}\label{section:Optimality_LiDAR_Bundle_Adjustment}

Without solving the full bundle adjustment problem \eqref{eq:lm}, we could still obtain the expected accuracy of \eqref{eq:lm} by analyzing the optimality of the relation graph $\mathbf{G}$. For a random edge $\mathbf{E}_k$, first and second elements are the $\mathbf{E}_{k}^{0}$-th pose and $\mathbf{E}_{k}^1$-th pose, respectively. The relative pose residual ($\boldsymbol{\epsilon}_k$) and covariance ($\mathbf{\Omega}_k$) between the $\mathbf{E}_{k}^{0}$-th pose and $\mathbf{E}_{k}^1$-th pose are approximated using the pair-wise registration formulation in \ref{appendix_covariance_reigstration}. The minimal eigen value of the covariance $\lambda^{min}_{k}=\lambda_{min}(\boldsymbol{\mathbf{\Omega}}_k)$ is used to measure the reliability of the edge $\mathbf{E}_k$. The optimality of $\mathbf{G}$ is first approximated by formulating the standard pose graph equation \eqref{eq:pgo},
\begin{equation}
  \underset{\mathbf{X}}{\mathrm{argmin}}~\sum_{k=1}^{l}\boldsymbol{\epsilon}_k^{\top}\mathbf{\Omega}_k\boldsymbol{\epsilon}_k. \label{eq:pgo}
\end{equation}
The information matrix $\boldsymbol{\varLambda}$ of \eqref{eq:pgo} is obtained as follows.
\begin{equation}
  \boldsymbol{\varLambda } =\sum_{k=1}^{l} \mathbf{J}^{\epsilon\top}_k\mathbf{\Omega}_k\mathbf{J}^{\epsilon}_k, \label{eq:pgo_hessian}
\end{equation}
where $\mathbf{J}^{\epsilon}_k$ is the Jacobian matrix of $\boldsymbol{\epsilon}_k$ with respect to $\mathbf{X}$. The information matrix $\boldsymbol{\varLambda }$ reveals the reliability of the poses to be estimated in bundle adjustment and also the optimality of the graph $\mathbf{G}$. The optimality of bundle adjustment for $\mathbf{G}$ is defined using the minimal eigenvalue of the information matrix: $\lambda_{min}(\boldsymbol{\varLambda })$ \citep{khanna2017scalable}.

\subsection{Schur Complement for State Marginalization}
The normal matrix $\mathbf{H}$ in \eqref{eq:lm} is of size $(6n)^2$, which can be impractically large and dense for large-scale LiDAR mapping tasks. A commonly used method to solve part of the linear system is applying Schur complement by reforming \eqref{eq:lm} as follows:
\begin{equation}
  \left[ \begin{matrix}
    \mathbf{H}_{aa}&		\mathbf{H}_{ab}\\
    \mathbf{H}_{ab}^{\top}&		\mathbf{H}_{bb}\\
  \end{matrix} \right] \left[ \begin{array}{c}
    \Delta \mathbf{X}_a\\
    \Delta \mathbf{X}_b\\
  \end{array} \right] =\left[ \begin{array}{c}
    \mathbf{y}_a\\
    \mathbf{y}_b\\
  \end{array} \right], 
\end{equation}
where $\Delta \mathbf{X}_a$ and $\Delta \mathbf{X}_b$ are the pose corrections to be solved (a small cluster of $\mathbf{X}$) and unsolved (the remaining large cluster of $\mathbf{X}$), respectively. $\Delta \mathbf{X}_a$ could be calculated using \eqref{eq:delta_x}.
\begin{equation}
  \left( \mathbf{H}_{aa}-\mathbf{H}_{ab}\mathbf{H}_{bb}^{-1}\mathbf{H}_{ab}^{\top} \right) \Delta \mathbf{X}_a=\mathbf{y}_a-\mathbf{H}_{ab}\mathbf{H}_{bb}^{-1}\mathbf{y}_b.\label{eq:delta_x}
\end{equation}

It is desirable that if $\mathbf{H}_{bb}$ is sparse for the inverse operation. The sparsity of  $\mathbf{H}_{bb}$ can be achieved by using edge selection, which will be detailed in Section \ref{section:goso}. Given an optimal update $\Delta \mathbf{X}_a$, the update for the remaining $\Delta \mathbf{X}_b$ is calculated by \eqref{eq:delta_y}.

\begin{equation}
  \mathbf{H}_{bb}\Delta \mathbf{X}_b=\mathbf{y}_b-\mathbf{H}_{ab}^{\top}\Delta \mathbf{X}_a,\label{eq:delta_y}
\end{equation}
which can be also regarded as conditional covariance of $\Delta \mathbf{X}_b$ given $\Delta \mathbf{X}_a$:
\begin{equation}
  \begin{split}
  \mathrm{Expectation}\left( \Delta \mathbf{X}_b | \Delta \mathbf{X}_a\right) &=\mathbf{H}_{bb}^{-1}\left( \mathbf{y}_b-\mathbf{H}_{ab}^{\top}\Delta \mathbf{X}_a\right),\\
  \mathrm{Covariance}\left( \Delta \mathbf{X}_b|\Delta \mathbf{X}_a\right) &=(\mathbf{H}_{bb}-\mathbf{H}_{ab}^{\top}\mathbf{H}_{aa}^{-1}\mathbf{H}_{ab}). \label{eq:marginalization}
  \end{split}
\end{equation}
Now, the state of $\mathbf{X}_a$ is marginalized and maintained as the prior information for $\mathbf{X}_b$. With this state marginalization scheme, we could solve large-scale LBA problems by separating the raw problem into small-scale sub-problems.

\begin{figure}[]
  \centering
  \includegraphics[width=0.48\textwidth]{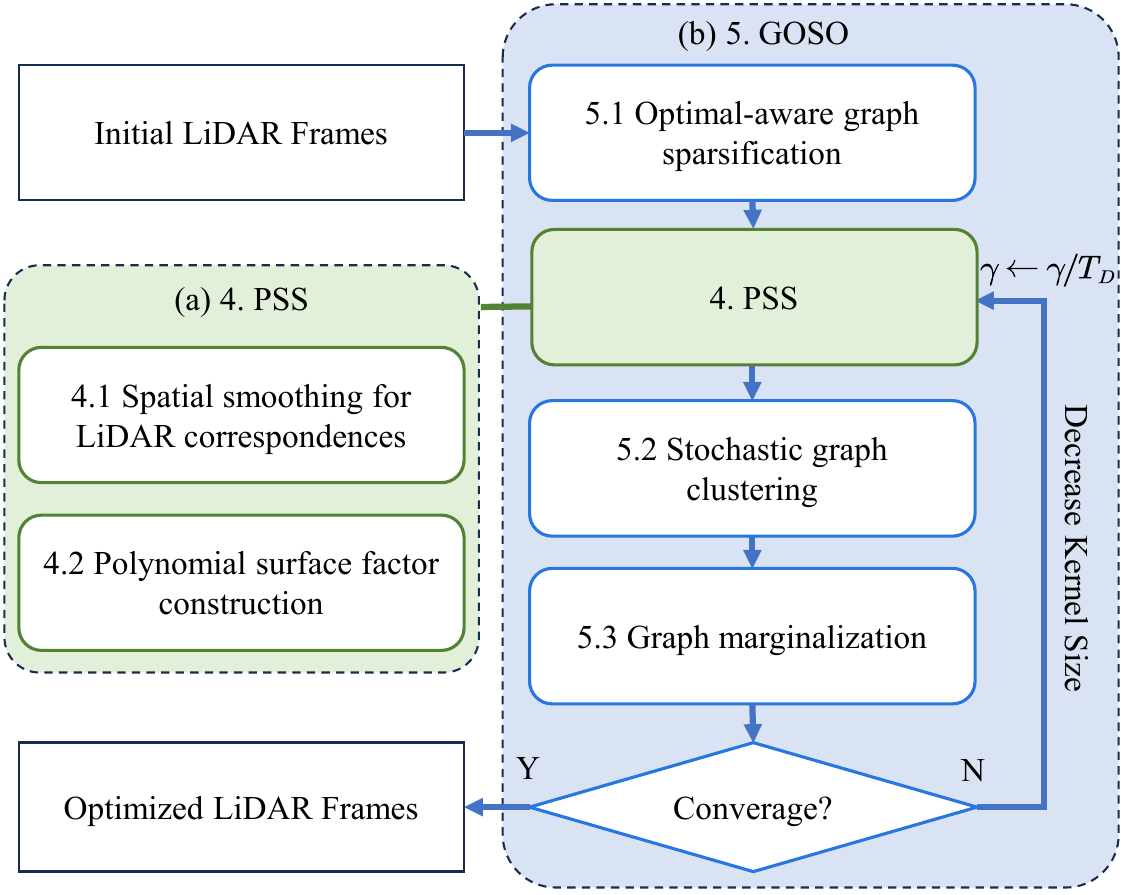}
  \caption{System overview of the proposed LBA approach. (a) Progressive Spatial Smoothing (PSS); (b) Graph Optimality-aware Stochastic Optimization (GOSO).}
  \label{fig:system}
  \vspace{-15pt}
\end{figure}

\section{System Overview}

To achieve a \textit{robust}, \textit{efficient}, and \textit{scalable} LBA, the proposed method shown in Fig.\ref{fig:system} includes two key modules, progressive spatial smoothing (PSS) detailed in Section \ref{section:pssba} and graph optimality-aware stochastic optimization (GOSO) detailed in Section \ref{section:goso}.

Taking the inaccurate initial LiDAR frames as input, GOSO first determines the relation graph of the LiDAR frames so that it achieves \textit{efficient} optimization (Section \ref{section:optimal_graph_sparsification}) according to the optimality of the graph. Second, PSS extracts the factors from the remaining edges in the sparse relation graph. It utilizes a spatial smoothing kernel to extract the LiDAR correspondences and a polynomial surface function to construct the factors for \textit{robust} pose correction. Then, GOSO separates the raw graph into subsets according to modularity~\citep{blondel2008fast} considering stochastic condition to avoid local minimal in the following optimization while maintaining good relationships in each subset (Section \ref{section:Stochastic_Graph_Clustering}). Finally, GOSO optimizes the LiDAR frames with graph marginalization in a successive manner to achieve a \textit{scalable} optimization (Section \ref{section:Graph_Marginalization}) for large-scale LBA. The above steps are repeated until the optimization converges or reaches the preset max iteration times (5 used in the experiment). After each iteration, the spatial kernel size $\gamma$ is decreased by $\gamma / T_D$. The detailed descriptions of the proposed method are as follows.
\section{Progressive Spatial Smoothing (PSS)\label{section:pssba}}

Extracting reliable LiDAR correspondences from LiDAR frames and constructing appropriate types of factors are crucial for robust pose adjustment. Noting that structural prior information can still be obtained even with imperfect initial poses, the proposed Progressive Spatial Smoothing (PSS) first extracts LiDAR correspondences and structural prior information through spatial smoothing. Subsequently, polynomial surface factors are constructed for the bundle adjustment process. It is important to note that only the correspondences between the remaining edges in the sparse relation graph (Section \ref{section:optimal_graph_sparsification}) are considered to ensure efficiency.

\subsection{Spatial Smoothing for LiDAR Correspondences}
\subsubsection{Smoothing Kernel Sampling} 
The input LiDAR frames are uniformly sampled using the voxel size of $\gamma$. Afterwards, the remaining points, $\{\hat{\mathbf{p}}^{W}_i, i \in \rm{\Psi}\}$, are treated as smoothing kernels. The initial point normals for each point in the noisy point clouds are calculated using Principal Component Analysis (PCA). For a given smoothing kernel, $\hat{\mathbf{p}}^{W}_i$, we define its neighboring points within the distance of $\gamma$ as $\{\hat{\mathbf{p}}^{W}_j,j\in\rm{\Phi}_i\}$. The inital point normal for the smoothing kernel $\hat{\mathbf{p}}^{W}_i$ and a neighborhood $\hat{\mathbf{p}}^{W}_j$ is respectively defined as $\breve{\mathbf{n}}_i$ and $\breve{\mathbf{n}}_j$.

Under the assumption of a continuous 3D environment, there is minimal variation in the normals across a local space. Consequently, we can obtain an optimal estimation of the normal, denoted as $\hat{\mathbf{n}}_i$ for the smoothing kernel $\hat{\mathbf{p}}^{W}_i$ by solving the objective function $\mathbf{G}(\hat{\mathbf{n}}_i)$ that constrains the change of normals with respect to neighborhood normals:
\begin{equation}
    \begin{split}
    \underset{\hat{\mathbf{n}}_i,|\hat{\mathbf{n}}_i|=1}{\mathrm{argmin}}\ \mathbf{G}(\hat{\mathbf{n}}_i) &= 1-\hat{\mathbf{n}}^{\top}_i\breve{\mathbf{n}}_i + \mu |\mathbf{D}(\hat{\mathbf{n}}_i)|_{0},\\
    \mathbf{D}(\hat{\mathbf{n}}_i)_j&= 1-\hat{\mathbf{n}}^{\top}_i\breve{\mathbf{n}}_j,
    \end{split}\label{eq:normal}
\end{equation}
where $\mathbf{D}(\hat{\mathbf{n}}_i)$ is the differential function for $\hat{\mathbf{n}}_i$ in the 3D space. The $L_0$ normalization counting the non-zero item is used here to eliminate the influence of outliers and preserve the original shape in the sharp regions \citep{xu2011image,sun2015denoising}. $\mu$ is the weight that balances the data term and smooth term. Equation \eqref{eq:normal} can be minimized using an auxiliary function which is detailed in \ref{appendix_auxiliary}.

Once we have obtained the optimal normal $\hat{\mathbf{n}}_i$ for the smoothing kernel $\hat{\mathbf{p}}^{W}_i$, the local tangent frame $F^S$ is established for $\hat{\mathbf{p}}^{W}_i$ by constructing three orthogonal axes:
\begin{equation}
\hat{\mathbf{n}}^i_2 = \hat{\mathbf{n}}_i, \hat{\mathbf{n}}^i_1 = [\hat{{n}}_{i,1},-\hat{{n}}_{i,0},0]^{\top}, \hat{\mathbf{n}}^i_0 = \hat{\mathbf{n}}^i_1 \times \hat{\mathbf{n}}_i.\\
\end{equation}
With these three axes, $\hat{\mathbf{M}}_i = [\hat{\mathbf{n}}^i_0,\hat{\mathbf{n}}^i_1,\hat{\mathbf{n}}^i_2]^{\top}$ is the matrix that tranforms the points from $F^W$ to $F^S$ as shown in Fig. \ref{fig:surface_fitting}.

\subsubsection{Weighted Surface Fitting}

\begin{figure}
  \centering
  \includegraphics[width=0.5\textwidth]{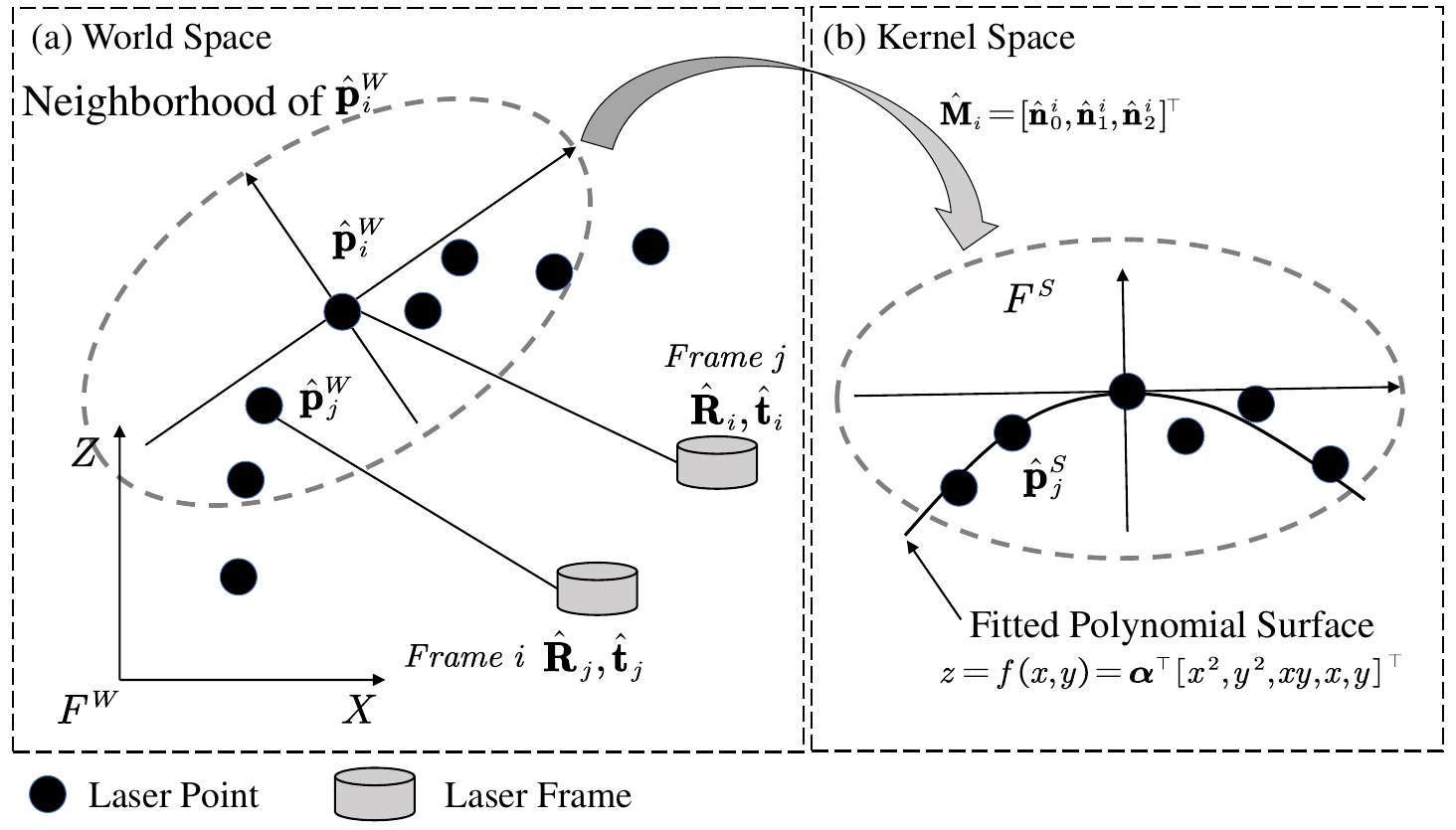}
  \caption{Surface fitting within a smoothing kernel. (a) Points in the world frame. (b) Points in the kernel space for polynomial fitting.}
  \label{fig:surface_fitting}
  \vspace{-15pt}
\end{figure}

For the smoothing kernel $\hat{\mathbf{p}}^{W}_i$, its neighborhood points $\{\hat{\mathbf{p}}^{W}_j,j\in\rm{\Phi}_i\}$ are projected to smoothing kernel's tangent space $F^S$:
\begin{equation}
    \left[ \hat{x}_{j}^{S},\hat{y}_{j}^{S},\hat{z}_{j}^{S} \right] ^{\top} =\hat{\mathbf{p}}_{j}^{S}=\hat{\mathbf{M}}_i\left( \hat{\mathbf{p}}_{j}^{W}-\hat{\mathbf{p}}_{i}^{W} \right).
\end{equation}

Let the paramters for the smoothing kernel $\hat{\mathbf{p}}^{W}_i$ be $\bm \alpha^i = [\alpha^i_0, \alpha^i_1,...,\alpha^i_5]^\top$, the smoothed coordinates of $\hat{\mathbf{p}}_{j}^{S}$ is $[\hat{x}_{j}^{S},\hat{y}_{j}^{S},f_i\left( \hat{x}_{j}^{S},\hat{y}_{j}^{S} \right)]^\top$, where $f_i\left( \hat{x}_{j}^{S},\hat{y}_{j}^{S} \right)$ is the polynomial function and calculated as follow:
\begin{equation}
    f_i\left( \hat{x}_{j}^{S},\hat{y}_{j}^{S} \right) = {\bm{\alpha}^i}^{\top}\left[ \left( \hat{x}_{j}^{S} \right) ^2,\left( \hat{y}_{j}^{S} \right) ^2,\hat{x}_{j}^{S}\hat{y}_{j}^{S},\hat{x}_{j}^{S},\hat{y}_{j}^{S} \right]^{\top}.
\end{equation}
Now the core problem is to robustly determine the optimal parameter $\bm \alpha^i$ using the projected noisy points $\{\hat{\mathbf{p}}^{S}_j,j\in\rm{\Phi}_i\}$. We find the best parameters $\bm \alpha^i$ using least-square estimation considering the Gaussian radial weight function $\mathrm{w}(d)$~\citep{alexa2003computing}:  
\begin{equation}
    \begin{split}
    \underset{\bm \alpha^i}{\mathrm{argmin}}\ \Sigma_{j\in\rm{\Phi}_i}& \left|\big(f_i\left( \hat{x}_{j}^{S},\hat{y}_{j}^{S} \right) - \hat{z}_{j}^{S}\big)\mathrm{w}(|\hat{\mathbf{p}}_{j}^{S}|)\right|^2,\\
    &\mathrm{w}(d)= e^{-d^2/\gamma^2}.
    \end{split}
\end{equation}
Then the fitted polynomial surface is obtained as shown in Fig. \ref{fig:surface_fitting}.

\subsubsection{Points Smoothing and Factor Association}
By replacing $\hat{z}_{j}^{S}$ with $f_i\left( \hat{x}_{j}^{S},\hat{y}_{j}^{S} \right)$ for each point in $\{\hat{\mathbf{p}}^{S}_j,j\in\rm{\Phi}_i\}$, we could obtain the smoothed point clouds. Moreover, the difference between $\hat{z}_{j}^{S}$ and $f_i\left( \hat{x}_{j}^{S},\hat{y}_{j}^{S} \right)$ is regarded as the error caused by the poses' error. Thus we associate $\{\hat{\mathbf{p}}^{S}_j,j\in\rm{\Phi}_i\}$ with $\mathbf{p}^{S}_i$, and use it for the following poses adjustment.
\subsection{Polynomial Surface Factor \label{sec:posescorrection}}

The difference between $\hat{z}_{j}^{S}$ and the fitted polynomial surface $f_i\left( \hat{x}_{j}^{S},\hat{y}_{j}^{S} \right)$ is regarded as the error $\sigma _{i,j}$ and written as:
\begin{equation}
    \sigma _{i,j}=f_i\left( \hat{x}_{j}^{S},\hat{y}_{j}^{S} \right) -\hat{z}_{j}^{S}. \label{eq:pss_residual}
\end{equation}
$\sigma _{i,j}$ is correlated with $i^{th}$ and $j^{th}$ poses, namely $[\hat{R}_i,\hat{t}_i]$ and $[\hat{R}_j,\hat{t}_j]$ as shown in Fig. \ref{fig:surface_fitting}. The Jacobian of $\sigma _{i,j}$ with respect to the $i^{th}$ pose can be derived using the chain rule in \ref{appendix_pss}.

\section{Graph Optimalily-Aware Stochastic Optimization (GOSO) \label{section:goso}}

\subsection{Optimality-aware Graph Sparsification \label{section:optimal_graph_sparsification}}
% how to construct the raw relation graph efficiently?
To build the relation graph $\mathbf{G}=(\mathbf{X},\mathbf{E})$ for the LBA problem, the LiDAR frames are first projected to the world frame according to the initial poses $\mathbf{\breve{X}}$. The projected point clouds are then voxelized using the voxel size of $\gamma$. In each voxel, it records the points from different frames. Any two frames that share more than $30\%$ overlaps of voxels will be used to establish an edge in $\mathbf{G}$. Meanwhile, the reliability of the edge between two frames will be calculated according to pair-wise point cloud registration (\ref{appendix_covariance_reigstration}).

The relation graph $\mathbf{G}$ usually contains dense edges due to the complex overlaps between the frames, making the solving procedure of bundle adjustment have low-efficiency. Thus, we sparsify $\mathbf{G}$ considering the optimal, namely, the minimal eigenvalue of the information matrix $\lambda_{min}(\boldsymbol{\varLambda })$ mentioned in Section \ref{section:Optimality_LiDAR_Bundle_Adjustment}. To make the raw dense relation graph $\mathbf{G}$ more sparse by only maintaining a subset of the raw edges $\mathbf{E}$ while keeping good optimality, it is formulated as an optimal design problem. To achieve this optimal design, we try to optimize the following objective function \eqref{eq:objective_function}.
\begin{equation}
  \begin{split}
  \mathrm{argmax}~\mathrm{F}_{set}(\mathbf{E}^{o})~ &s.t.~ \mathbf{E}^{o}\subset \mathbf{E},\vert \mathbf{E}^{o} \vert \leqslant \mathbf{N}^{o},\\
  \mathrm{F}_{set}(\circ)&=\lambda_{min}(\boldsymbol{\varLambda}(\circ)), \label{eq:objective_function}
  \end{split}
\end{equation}
where $\mathbf{N}^{o}$ is the desired number of remaining edges $\mathbf{E}^{o}$ in the optimal designed graph $\mathbf{G}^{o}=(\mathbf{X},\mathbf{E}^{o})$. Equation \eqref{eq:objective_function} is an NP-hard problem, while it can still be approximately solved efficiently using stochastic-greedy heuristic searching \citep{khanna2017scalable}. The stochastic-greedy algorithm starts with an empty set. Then, at each step, it selects one element from a random set with a size of $\mathrm{log}(1/\epsilon) \vert \mathbf{E} \vert$ from the remaining edge set, which gains most of the objective function $\mathrm{F}_{set}(\mathbf{E}^{o})$. $\epsilon$ is the decay factor. The iterative process stops until it gets $\mathbf{N}^{o}$ edges. In the real experiment, $\mathbf{N}^{o}$ can be set according to a certain percentage (e.g., $20\%$) of the size of raw edges. The details of the stochastic-greedy heuristic searching can be found in algorithm \ref{algo:Stochastic-Greedy}.  

\begin{algorithm}
  \DontPrintSemicolon
  \SetAlgoLined
  \KwIn { $\mathbf{E}$, $\epsilon$, $\mathbf{N}^{o}$}
  \KwOut { Selected edges $\mathbf{E}^{o}$}
  Initialization of the set  $\mathbf{E}^{o} \gets \oslash $;\;
  
  \While{ $\mathbf{E}^{o} < \mathbf{N}^{o} $}{
      $\mathbf{S} \gets $ random sampling $\mathrm{log}(1/\epsilon) \vert \mathbf{E} \vert$ items from $\mathbf{E}$;\;
      \ForEach{${\mathbf{S}}_v \in \mathbf{S}  $}{
      Compute $\boldsymbol{\varLambda}_v = \mathbf{J}^{\epsilon\top}_v\mathbf{J}^{\epsilon}_v\;; // See~Section~\ref{section:Optimality_LiDAR_Bundle_Adjustment}$
      }
      $v^{*}\gets \mathrm{argmax}_{v\in \mathbf{S}} \lambda_{min}[\boldsymbol{\varLambda}(\mathbf{E}^{o})+\boldsymbol{\varLambda}_v] $;\;
      $\boldsymbol{\varLambda}(\mathbf{E}^{o}) \gets \boldsymbol{\varLambda}(\mathbf{E}^{o})+\boldsymbol{\varLambda}_{v*}$;\;
      $\mathbf{E}^{o} \gets \mathbf{E}^{o} \bigcup \mathbf{S}_{v*}$; $\mathbf{E} \gets \mathbf{E} \backslash  \mathbf{S}_{v*}$. \;
      
  }
  \caption{Stochastic-Greedy for Optimality-aware Graph Sparsification \label{algo:Stochastic-Greedy}}
\end{algorithm}

\begin{thm}[\cite{mirzasoleiman2015lazier}]
  The set function $\mathrm{F}_{set}$ is submodular and monotone increasing.
\end{thm}

\begin{thm}[\cite{nemhauser1978analysis}]
  Let $\mathbf{E}^{o*}$ and $\mathbf{E}^{o\#}$ be the global optimal and the optimal obtained using the stochastic-greedy heuristic, then 
  \begin{equation}
    \begin{split}
      \frac{\mathrm{F}_{set}(\mathbf{E}^{o*})-\mathrm{F}_{set}(\mathbf{E}^{o\#})}{\mathrm{F}_{set}(\mathbf{E}^{o*})} \leq \left(\frac{\mathbf{N}^{o}-1}{\mathbf{N}^{o}}\right)^{\mathbf{N}^{o}}+\epsilon \leq \frac{1}{e} + \epsilon.\label{eq:worst_bound}
    \end{split}
  \end{equation}
  Equation \eqref{eq:worst_bound} elicits a lower bound for the stochastic-greedy heuristic results. In practice, the stochastic-greedy heuristics will achieve better results than the lower bound.
\end{thm}

\begin{thm}[\cite{nemhauser1978analysis}]
The time complexity of the stochastic-greedy searching is $O(\mathrm{log}(1/\epsilon) \vert \mathbf{E} \vert)$.
\end{thm}

\subsection{Stochastic Graph Clustering \label{section:Stochastic_Graph_Clustering}}

% 利用不确定度，度量边的好坏。

After the graph sparsification, the large number of nodes in the relation graph $\mathbf{G}$ still limits the efficiency and scalability of the LBA problem. Thus, we propose a stochastic graph clustering to split the relation graph $\mathbf{G}$ into subsets considering the following requirements. First, the stochastic graph clustering is intended to partition the nodes into small independent clusters and the nodes in each cluster share a tight relationship, so the LBA \eqref{eq:bundle_adjustment} could be solved efficiently. Second, the stochastic graph clustering should be randomized to avoid local minimum in the optimization.

The proposed stochastic graph clustering operates on the relation graph $\mathbf{G}$, where the weight of each edge $\mathbf{E}_k$ is measured using the minimal eigenvalue of the covariance $\lambda^{min}_{k}$, which is introduced in Section \ref{section:Optimality_LiDAR_Bundle_Adjustment}. At the beginning, each node formulates an individual cluster. Next, the nodes that share an edge will be merged as a new cluster. In order to make the nodes in each cluster share a tight relationship, an optimal clustering is to be found to maximize the modularity $\mathbf{Q}$ \eqref{eq:modularity}~\citep{blondel2008fast}.

\begin{equation}
  \begin{split}
  \mathbf{Q}=\frac{1}{2\mathrm{s}}\sum_{\mathbf{E}_k\in \mathbf{E}}\delta(\mathbf{E}^0_k,\mathbf{E}^1_k)\left(\lambda^{min}_{k}-\frac{\mathrm{w}(\mathbf{E}^0_k)\mathrm{w}(\mathbf{E}^1_k)}{2\mathrm{s}}\right), \label{eq:modularity}
  \end{split}
\end{equation}
where $\mathrm{s}=\sum_{\mathbf{E}_k\in \mathbf{E}}\lambda^{min}_{k}$ is the total sum of the edge weights. $\mathrm{w}(\mathbf{E}^0_k)$ is the sum of weights of edges incident to the $\mathbf{E}^0_k$-th node. $\delta(\mathbf{E}^0_k,\mathbf{E}^1_k)=1$ if the $\mathbf{E}^0_k$-th node and the $\mathbf{E}^1_k$-th node belong to one cluster, otherwise $\delta(\mathbf{E}^0_k,\mathbf{E}^1_k)=0$.

\begin{figure}
  \centering
  \includegraphics[width=\linewidth]{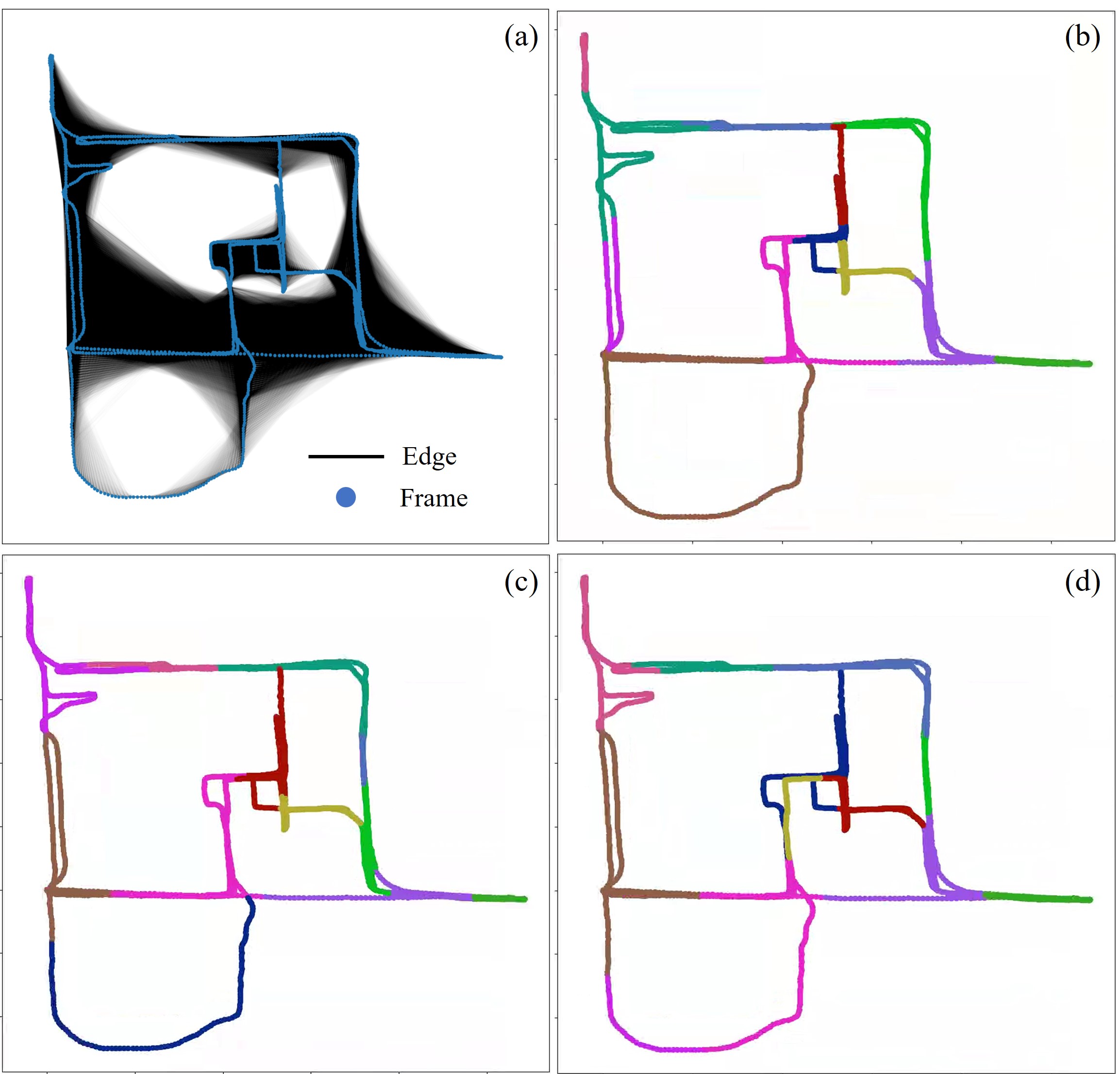}
  \caption{Example of stochastic clustering. (a) The sparsified graph of an example bundle adjustment problem; (b),(c), and (d) illustrates the different stochastic clustering results from three optimization steps. The random colors illustrate different clusters. }
  \label{fig:stochastic_clustering}
  \vspace{-15pt}
\end{figure}

\begin{figure*}
  \centering
  \includegraphics[width=\linewidth]{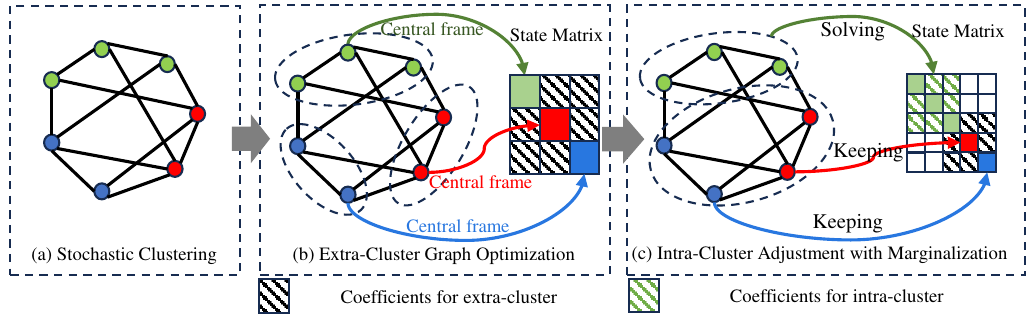}
  \caption{Graph marginalization.}
  \label{fig:graph_marginalization}
  \vspace{-15pt}
\end{figure*}

Finding the optimal clustering that maximizes the modularity is an NP-hard problem. Louvain's algorithm \citep{blondel2008fast}, greedily merging two clusters that give the maximum increase of the modularity is a standard and efficient strategy to solve the problem \citep{blondel2008fast}. However, Louvain's algorithm lacks randomization due to the greedy strategy. Inspired by \cite{zhou2020stochastic}, we randomly merge clusters according to a probability distribution based on modularity increments. In other words, the larger the modularity increment, the more likely the two clusters are to merge. The maximum size of the items in each cluster is set to $T_{\mathrm{cluster}}$ (e.g., 300 is used in the experiment). The random merging is terminated when the modularity is not increasing. Figure \ref{fig:stochastic_clustering} illustrates the different clustering results of a relation graph. The proposed stochastic graph clustering achieves different but meaningful results in each iteration to achieve high accuracy while avoiding local minimum.

\subsection{Graph Marginalization \label{section:Graph_Marginalization}}
The proposed graph marginalization is used to solve the raw large-scale LBA problem by dividing the original problem into subproblems as illustrated in Figure \ref{fig:graph_marginalization}. Taking the stochastic clustering results as input (Figure. \ref{fig:graph_marginalization} (a)), the proposed graph marginalization first conducts the extra-cluster graph optimation (Figure. \ref{fig:graph_marginalization} (b)) treating each cluster as a rigid body, which only optimizes the relative poses between clusters. Then, intra-cluster adjustment with marginalization (Figure. \ref{fig:graph_marginalization} (c)) is utilized to adjust the poses within each cluster. The global information, namely the relative poses between clusters is considered as the prior marginalized constraints in this step. These two main steps are iterated multiple times, as detailed below.

\begin{rmk}
  Dividing the raw large-scale bundle adjustment problem into subproblems is a common strategy for the image or LBA. For example, the hierarchical LBA \citep{liu2023large} split the raw problem into subproblems according to the length of the successive frames, which does not consider the spatial relationship between the frames. Our problem-dividing strategy is inspired by the large-scale image bundle adjustment \citep{jiang2024efficient} and the reduced camera system \citep{demmel2021square}, which consider the spatial relationships between the images using the initial values of the cameras leading to better accuracy of the final results.
\end{rmk}

\subsubsection{Extra-Cluster Graph Optimization}
The relative poses between clusters are first optimized and extracted to guide the subsequent local adjustment. In this step, the center frame in each cluster is selected, then the relative poses of other frames respective to the center frame in the cluster are fixed. In this manner, each cluster is treated as a rigid body, and only the selected central poses will be optimized considering the constraints between clusters at this extra-cluster graph optimization stage. Thus, the number of poses to be optimized in this stage is equal to the number of the clusters as shown in Figure. \ref{fig:graph_marginalization} (b), so the size of the state matrix is acceptable for the optimization. The constraints involved in this stage are the PSS factors between clusters.

\subsubsection{Intra-Cluster Adjustment with Marginalization}
The poses in each cluster are optimized with the global guidance from the extra-cluster graph optimization in this step. Taking Figure \ref{fig:graph_marginalization} (c) as an example, the poses marked as green dots belong to one cluster and will be optimized. The global prior information is obtained by marginalization of the states from other clusters (marked as red and blue dots) for the center frame in the green cluster. More specifically, the global prior information for the current cluster is obtained using the equation \eqref{eq:marginalization}. The states of the red and blue dots are marginalized and maintained as the prior information for the states of green dots in Figure \ref{fig:graph_marginalization} (c). It should be noted that the proposed intra-cluster adjustments are operated in parallel for each cluster. 

\section{Experiments}\label{section:experiments}

\subsection{Implementation}

We implement the proposed PSS-GOSO in C++. The initial kernel size $\gamma$ is set to 3m. The decreasing factor $T_D$ is set to 1.4. The desired number of edges $\mathbf{N}^{o}$ is set according to $20\%$ of the raw edges. All algorithms are evaluated on a computer with an Intel Core I9-I2900 CPU.

\subsection{Evaluation on Ground-based Benchmark Dataset}

\begin{figure*}
  \centering
  \includegraphics[width=\linewidth]{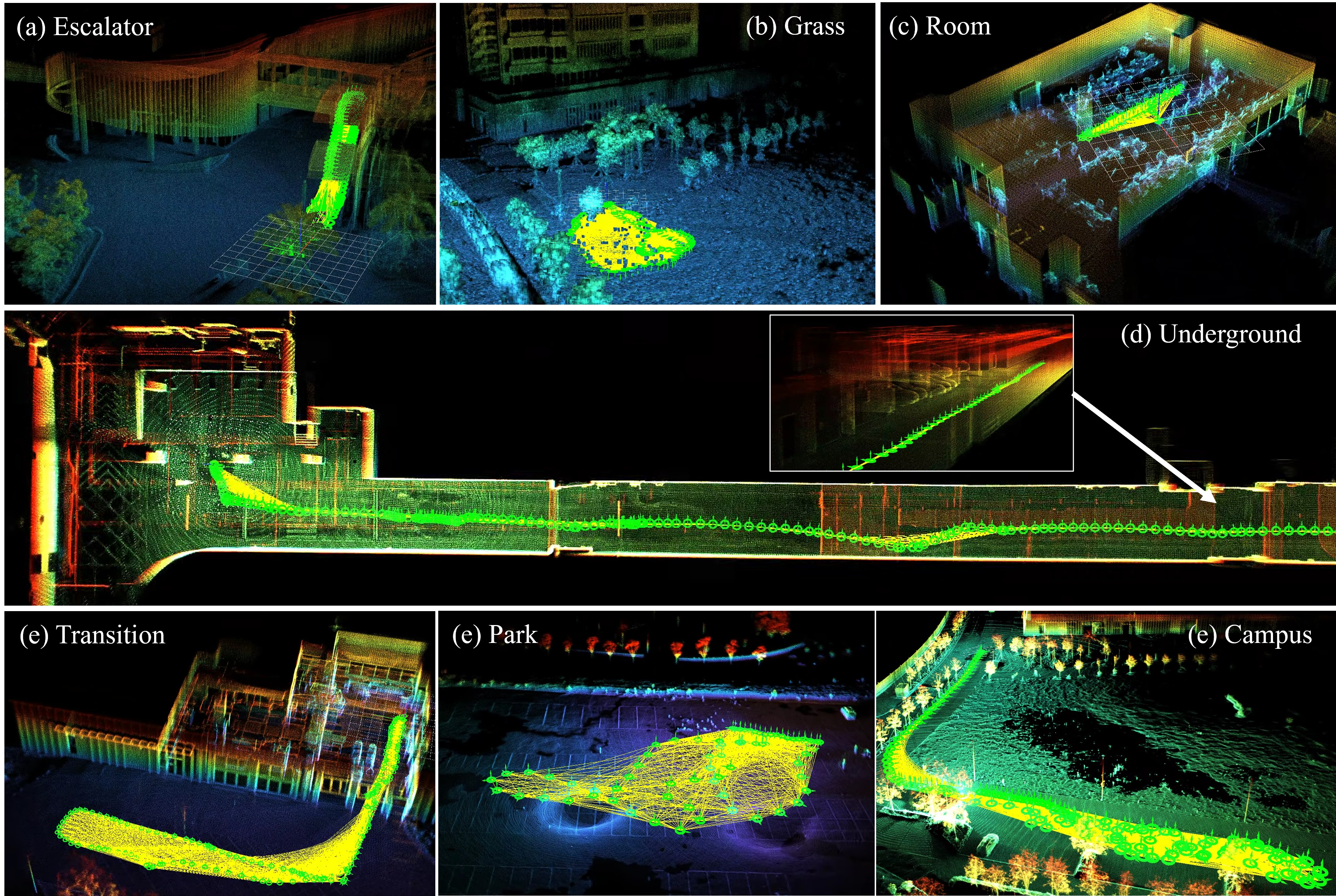}
  \caption{The point clouds and dense graphs in FusionPortable~\citep{wei2024fusionportablev2} dataset. Edges of the graph are marked as yellow lines. The nodes of the graph are marked as green axes.}
  \label{fig:results_fusionportable}
  \vspace{-15pt}
\end{figure*}

\begin{figure*}
  \centering
  \includegraphics[width=\linewidth]{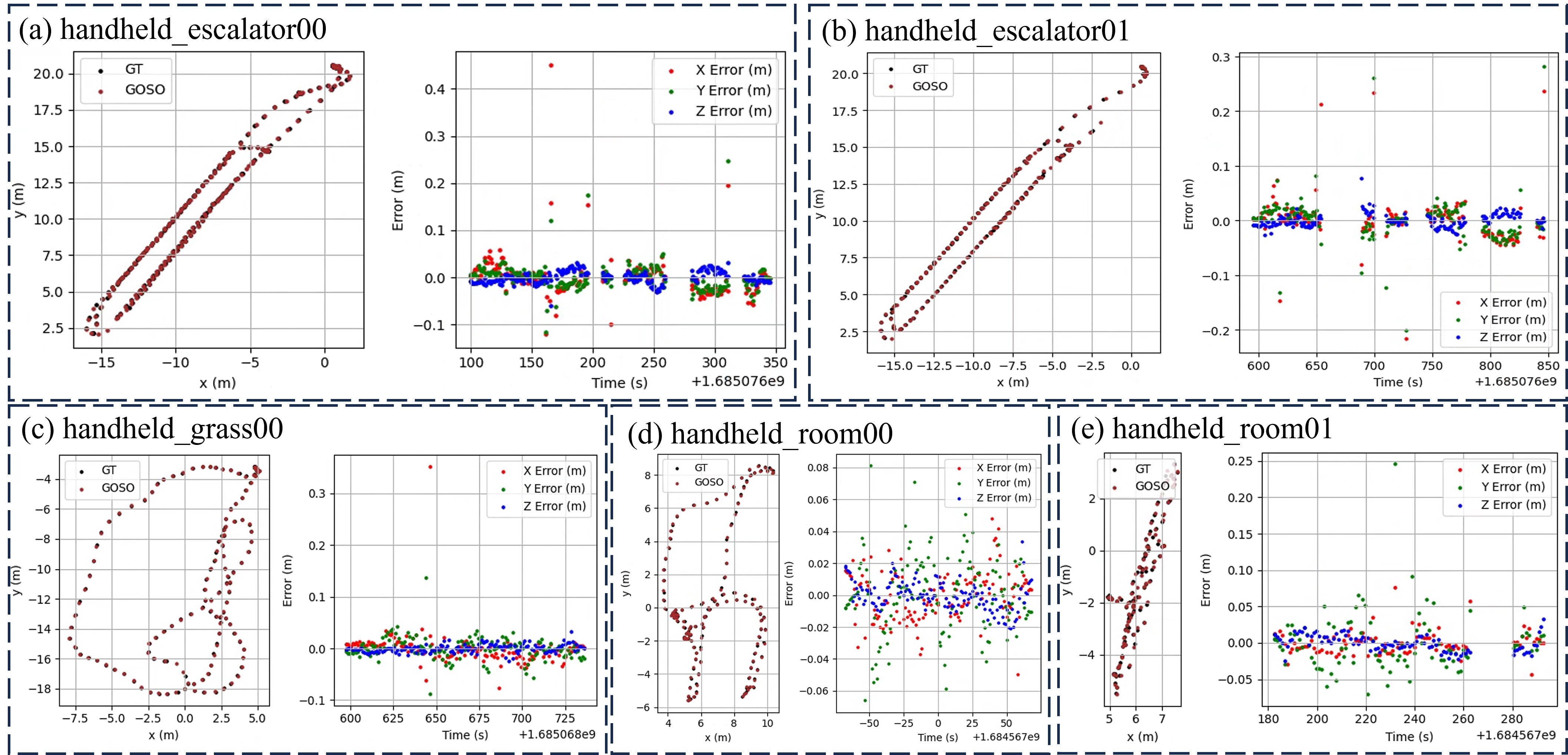}
  \caption{Absolute Position Error (APE) distribution of PSS-GOSO in the handheld sequences of FusionPortable~\citep{wei2024fusionportablev2} dataset.}
  \label{fig:APE_FUNSIONPORTABLE_HANDHELD}
  \vspace{-15pt}
\end{figure*}

\begin{figure}
  \centering
  \includegraphics[width=\linewidth]{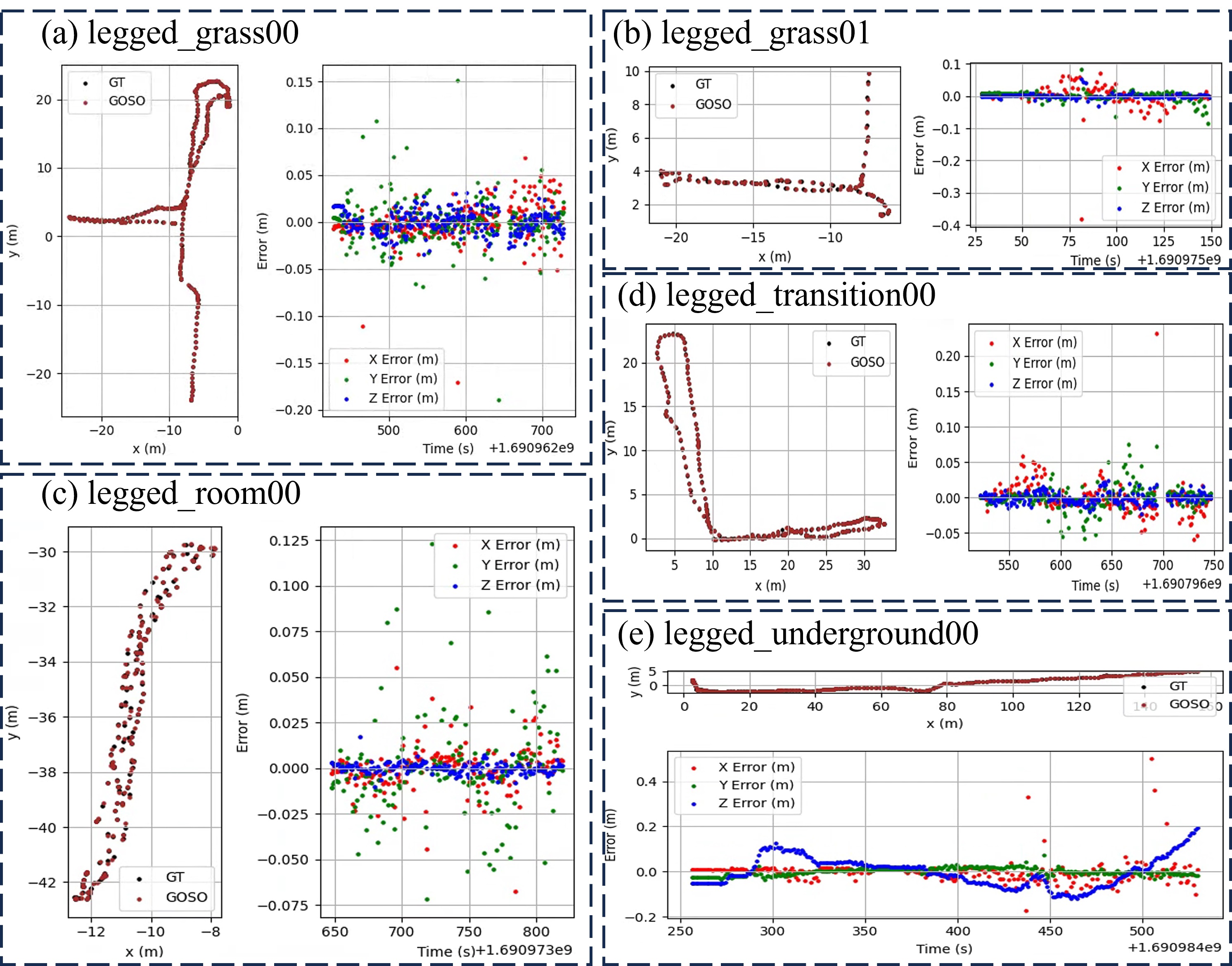}
  \caption{Absolute Position Error (APE) distribution of PSS-GOSO in the legged sequences of FusionPortable~\citep{wei2024fusionportablev2} dataset.}
  \label{fig:APE_FUNSIONPORTABLE_LEGGED}
  \vspace{-15pt}
\end{figure}

\begin{figure}
  \centering
  \includegraphics[width=\linewidth]{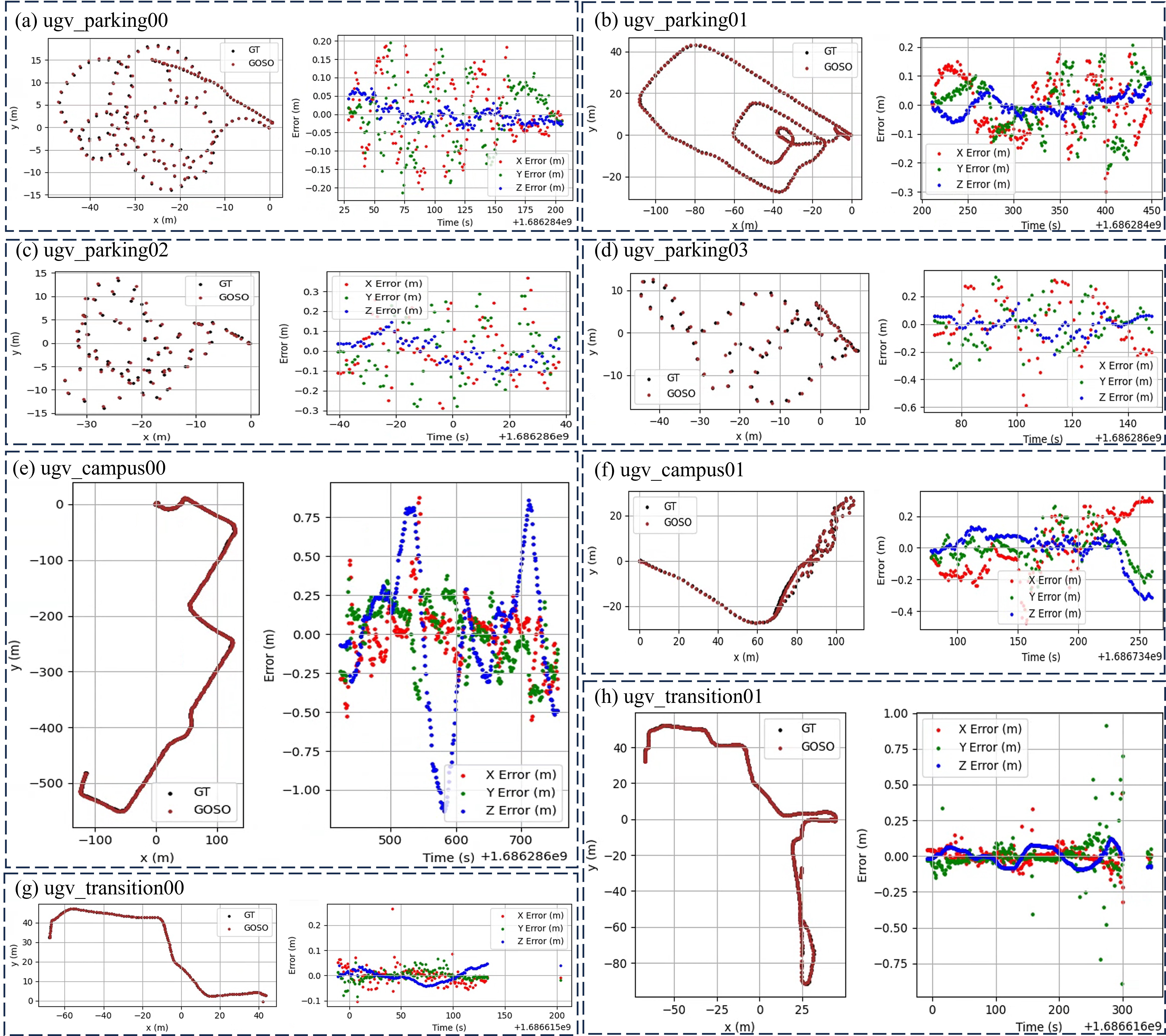}
  \caption{Absolute Position Error (APE) distribution of PSS-GOSO in the UGV sequences of FusionPortable~\citep{wei2024fusionportablev2} dataset.}
  \label{fig:APE_FUNSIONPORTABLE_UGV}
  \vspace{-15pt}
\end{figure}

We first evaluate the proposed PSS-GOSO on the ground-based benchmark dataset, named FusionPortable~\citep{wei2024fusionportablev2}. The FusionPortable contained diverse platforms and diverse environments, covering the main scenes that need accurate 3D mapping. Using the initial poses obtained by Fast-lio~\citep{xu2022fast} and manual loop closure, the dense graph and point clouds generated by PSS-GOSO in the diverse environments are demonstrated in Fig. \ref{fig:results_fusionportable}. We mainly focus on the handheld, legged, and UGV platforms which often operate in complex environments without GNSS. We compare the proposed method with the existing pose-graph-based method \citep{koide2019portable}. The APE distribution of the proposed method in the handheld, legged, and UGV sequences are plotted in Fig. \ref{fig:APE_FUNSIONPORTABLE_HANDHELD}, Fig. \ref{fig:APE_FUNSIONPORTABLE_LEGGED}, and Fig. \ref{fig:APE_FUNSIONPORTABLE_UGV}, respectively. The APE of different methods are listed in Table.~\ref{tab:APE_FUSIONPORTABLE}, which demonstrate that the proposed PSS-GOSO improved the initial pose accuracies. However, the existing pose-graph-based method ~\citep{koide2019portable} may suffer from inaccurate pair-wise registration thus achieving less accuracy.

\begin{table}[]
  \caption {Absolute Position Error (APE) of different methods in FusionPortable~\citep{wei2024fusionportablev2} dataset [m]. The best results are in \textbf{BOLD}.}
  \label{tab:APE_FUSIONPORTABLE}
  \centering
  \scalebox{0.8}{
  \begin{tabular}{c|c|cccc}\hline
                    & Sequence              & \makecell[c]{Initial \\ (SLAM + Loop) ~\citep{xu2022fast}} &PSS-GOSO    & \makecell[c]{Pose-graph\\\citep{koide2019portable}} &\makecell[c]{HBA\\ \cite{liu2023large}}\\ \hline
  \multirow{5}{*}{\rotatebox{90}{Handheld}} & escalator00 &    0.08     &      \textbf{0.05}      &     0.06   & 0.06\\
                            & escalator01 &    0.07     &      \textbf{0.06}       &    0.07    & 0.07 \\
                            & grass00     &    0.05     &      \textbf{0.04}       &    0.05    & 0.05\\
                            & room00      &    0.06     &      \textbf{0.03}       &    0.05    & 0.05\\
                            & room01      &    0.08     &      \textbf{0.04}       &    0.07    & 0.08\\ \hline
  \multirow{5}{*}{\rotatebox{90}{Legged}}   & grass00       &  0.07   &      \textbf{0.04}      &    0.06    & 0.06\\
                            & grass01       &   0.09      &      \textbf{0.05}       &    0.08   & 0.08\\
                            & room00        &   0.05      &      \textbf{0.04}       &    0.05    & 0.05\\
                            & transition00  &   0.05      &      \textbf{0.03}       &    0.05    & 0.04\\
                            & underground00 &   0.11      &      \textbf{0.09}       &    0.15    & 0.12\\ \hline
  \multirow{8}{*}{\rotatebox{90}{UGV}}      & parking00        &   0.29      &      \textbf{0.12}       &   0.23     & 0.26\\
                            & parking01        &   0.94      &     \textbf{0.14}        &    0.27    & 0.34\\
                            & parking02        &   0.42      &     \textbf{0.20}        &    0.41    & 0.40\\
                            & parking03        &   0.54      &     \textbf{0.25}        &    0.49    & 0.51\\
                            & campus00         &   1.31      &     \textbf{0.50}        &    1.01    & 1.20\\
                            & campus01         &   0.53      &     \textbf{0.23}        &    0.45    & 0.43\\
                            & transition00     &   0.14      &     \textbf{0.05}        &    0.09    & 0.07\\
                            & transition01     &   0.20      &     \textbf{0.16}        &    0.22    & 0.19\\ \hline
  \end{tabular}
  }
  \vspace{-15pt}
\end{table}

\subsection{Evaluation on Air-based Benchmark Dataset}

\begin{figure}
  \centering
  \includegraphics[width=\linewidth]{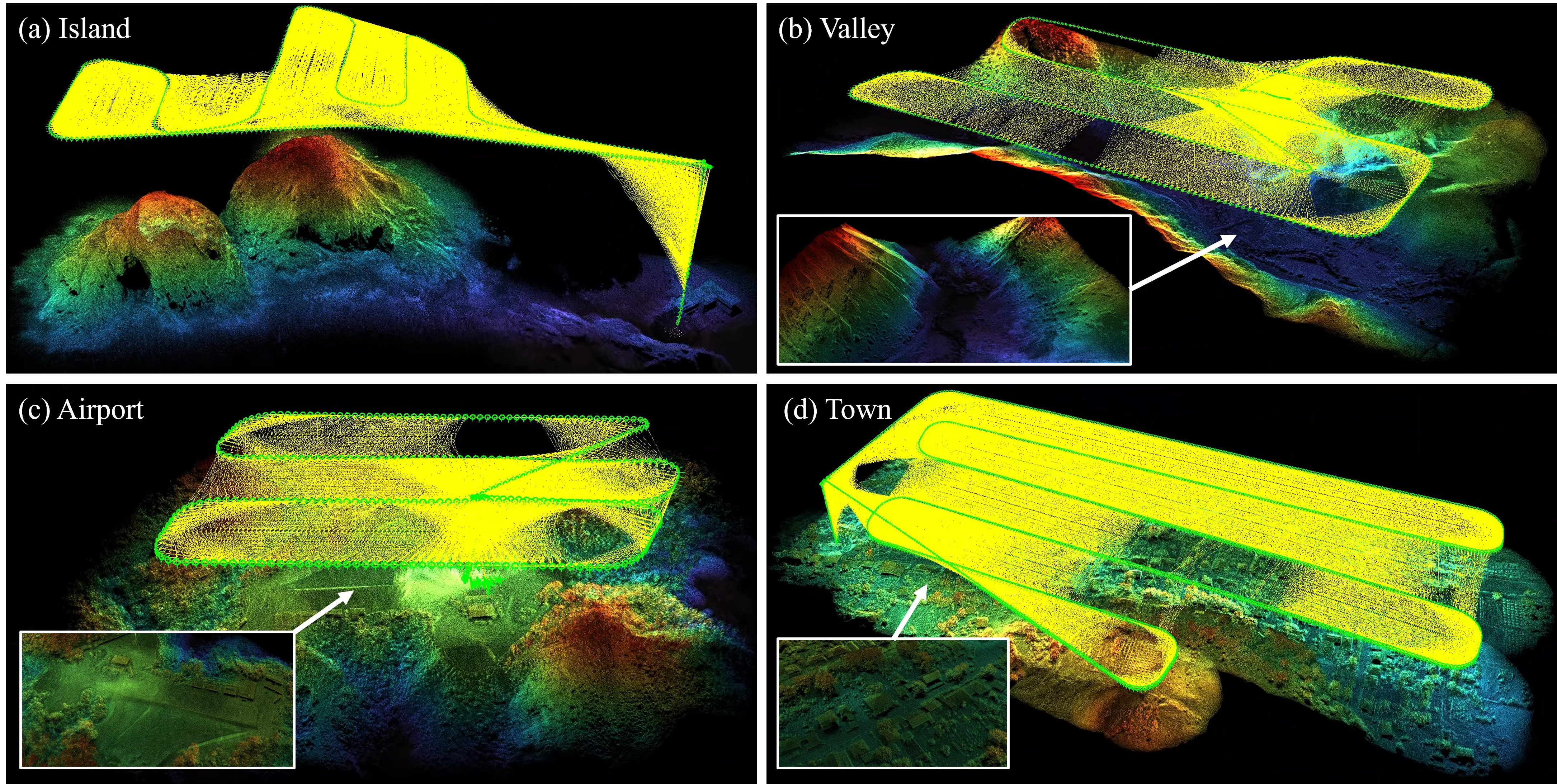}
  \caption{The point clouds and dense graphs in MARS-LVIG~\citep{li2024mars} dataset. Edges of the graph are marked as yellow lines. The nodes of the graph are marked as green axes.}
  \label{fig:results_marsuav}
  \vspace{-15pt}
\end{figure}

\begin{figure*}
  \centering
  \includegraphics[width=\linewidth]{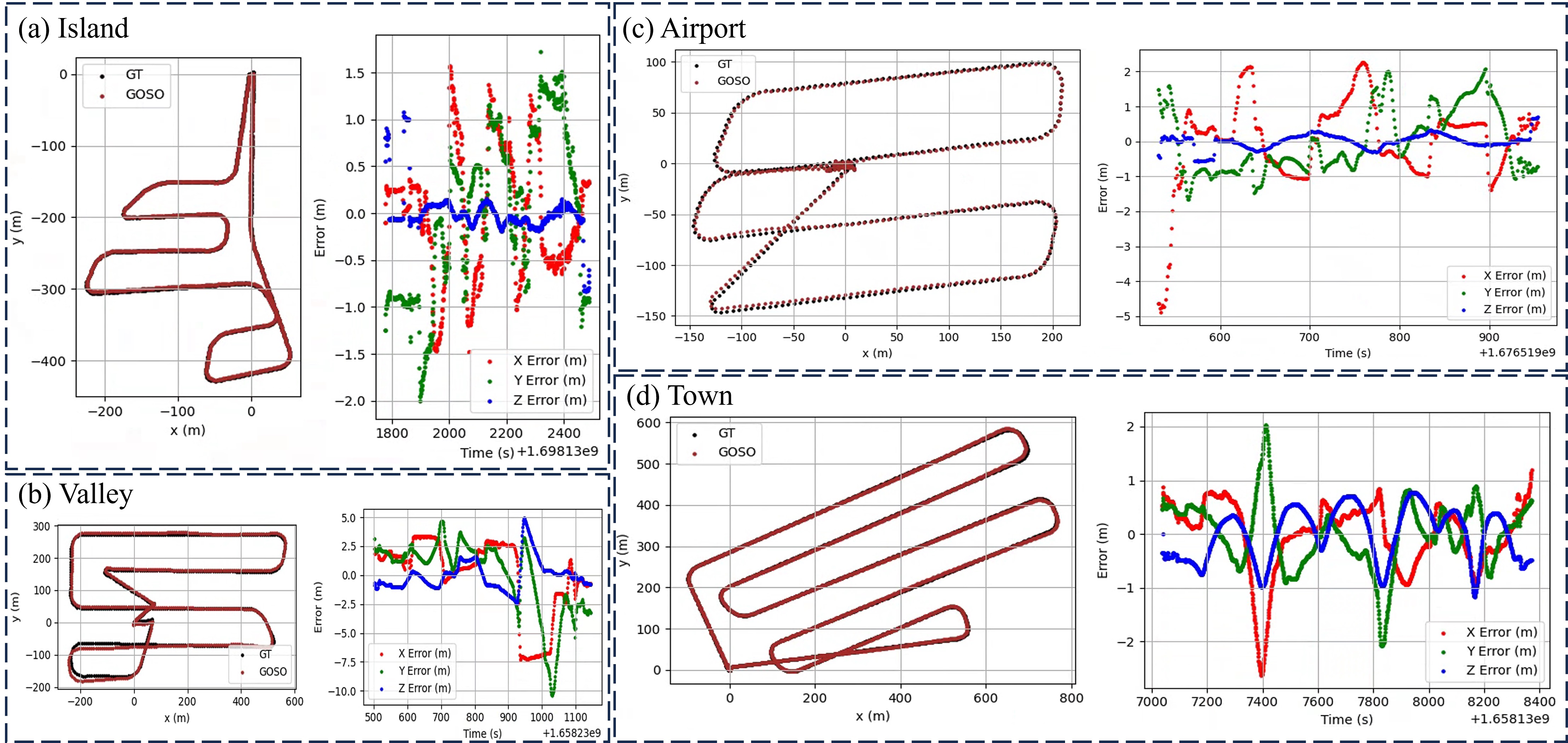}
  \caption{Absolute Position Error (APE) distribution of PSS-GOSO in MARS-LVIG~\citep{li2024mars} dataset.}
  \label{fig:APE_MARS_UAV}
  \vspace{-15pt}
\end{figure*}

Aerial LiDAR point cloud quality is crucial for photogrammetry. However, GNSS is very vulnerable to accidental or malicious radio frequency interruptions (jamming) or fake signals (spoofing) \cite{harris2021military,keetha2023anyloc}. We assess the performance of our proposed PSS-GOSO without GNSS singals using the MARS-LVIG benchmark dataset \citep{li2024mars}, which includes diverse environments such as islands, valleys, airports, and towns. Initial pose estimates were obtained using Fast-LIO \citep{xu2022fast} and manual loop closure. The dense graph and point clouds generated by PSS-GOSO in the diverse environments are demonstrated in Fig. \ref{fig:results_marsuav}. The APE distribution of the proposed method in the UAV sequences is plotted in Fig.~\ref{fig:APE_MARS_UAV}.  The APE of different methods is listed in Table.~\ref{tab:APE_MARS_UAV}. The results demonstrate that PSS-GOSO significantly improves accuracy compared to other methods. For instance, in the island sequence, PSS-GOSO reduces the APE to 1.14 meters, improving on the initial 2.37 meters and surpassing both the pose-graph-based method \citep{koide2019portable} and the HBA method \citep{liu2023large}. In the valley sequence, PSS-GOSO achieves an APE of 4.75 meters, a substantial reduction from the initial 11.99 meters and better than the pose-graph method’s 7.37 meters and HBA’s 6.40 meters. For the airport sequence, PSS-GOSO’s APE of 1.60 meters outperforms the initial 3.06 meters and the results from other methods. In the town sequence, PSS-GOSO achieves a notable APE improvement compared to the initial 2.37 meters, with better results than the pose-graph method and HBA. These results highlight PSS-GOSO’s effectiveness in enhancing the precision of aerial LiDAR point clouds across various challenging environments.

\begin{table}[]
  \caption {Absolute Position Error (APE) of different methods in MARS-LVIG~\citep{li2024mars} dataset [m]. The best results are in \textbf{BOLD}.}
  \label{tab:APE_MARS_UAV}
  \centering
  \scalebox{0.9}{
  \begin{tabular}{c|c|cccc}\hline
                    & Sequence              & \makecell[c]{Initial  \\(SLAM + Loop)\citep{xu2022fast}} &PSS-GOSO    & \makecell[c]{Pose-graph\\~\citep{koide2019portable}} & \makecell[c]{HBA\\\citep{liu2023large}}\\ \hline
  \multirow{5}{*}{\rotatebox{90}{UAV}} & Island &   2.37   &    \textbf{1.14}     &   1.98   & 2.24\\
                            & Valley &   11.99   &    \textbf{4.75}      &   7.37  & 6.40\\
                            & Airport      &    3.06  &     \textbf{1.60}     &    2.13 & 1.91\\
                            & Town &  2.37    &    \textbf{1.02}     &  1.85   & 1.80\\ \hline
  \end{tabular}
  }
\end{table}

\subsection{Evaluation on Proprietary Large-scale Port Dataset}

\begin{figure}
  \centering
  \includegraphics[width=\linewidth]{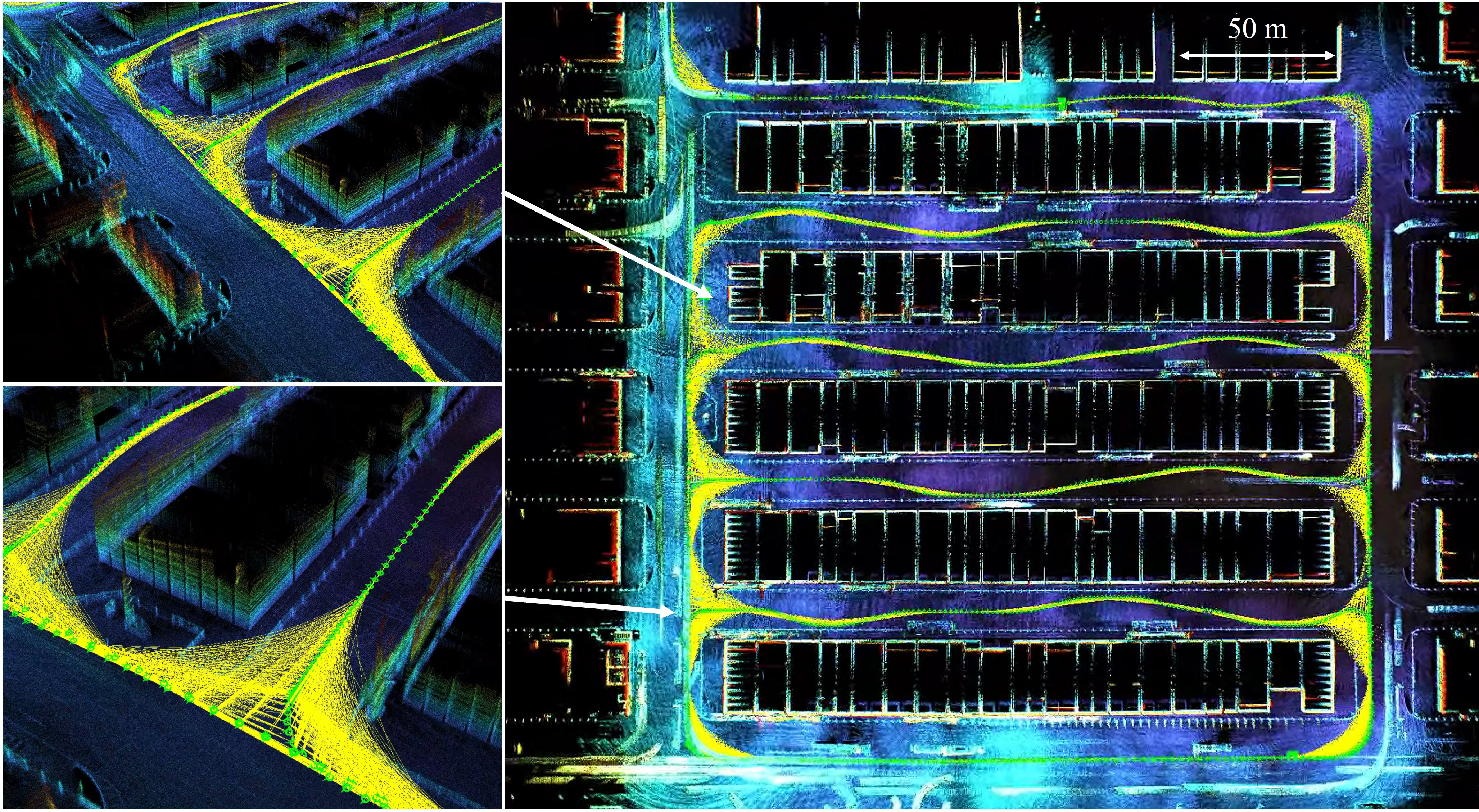}
  \caption{The point clouds and dense graphs in the proprietary large-scale port dataset. Edges of the graph are marked as yellow lines. The nodes of the graph are marked as green axes.}
  \label{fig:results_port}
\end{figure}

\begin{figure}
  \centering
  \includegraphics[width=\linewidth]{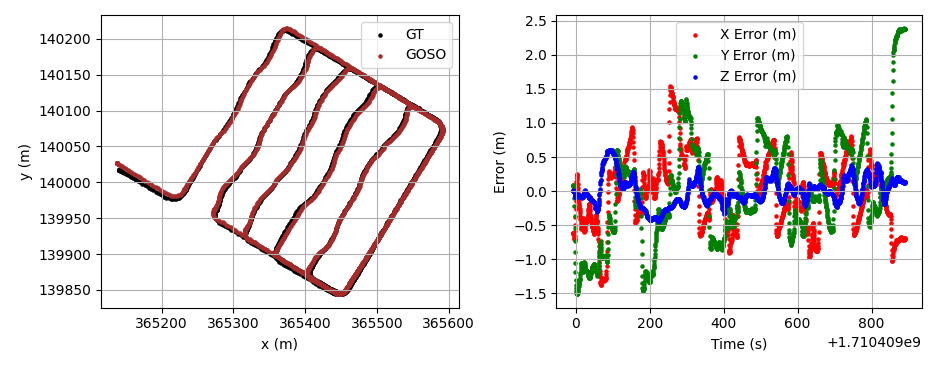}
  \caption{Absolute Position Error (APE) distribution of PSS-GOSO in the large-scale port dataset.}
  \label{fig:APE_PORT}
  \vspace{-15pt}
\end{figure}

The 3D prior map is essential for automatic robot delivery in large-scale port environments, where the complexity of the scene and the high volume of data present significant challenges. We conduct a comprehensive evaluation using data collected from an Ouster OS1-64 laser scanner across a $400 \text{m} \times 400 \text{m}$ port area, supplemented by RTK-GNSS data for ground truth. Initial pose estimates are obtained using Fast-LIO and manual loop closure. The dense graph and point clouds generated by PSS-GOSO are illustrated in Fig. \ref{fig:results_port}, and the Absolute Pose Error (APE) distribution is shown in Fig. \ref{fig:APE_PORT}. The experimental results, summarized in Table \ref{tab:APE_port}, reveal that PSS-GOSO achieves an APE of 1.31 meters for the port UGV sequence. This performance significantly surpasses the initial APE of 8.63 meters, the APE of 4.33 meters from the pose-graph method \citep{koide2019portable}, and the 3.31 meters APE from the hierarchical bundle adjustment (HBA) method \citep{liu2023large}. These results highlight PSS-GOSO's superior accuracy and effectiveness in handling large-scale, complex environments, demonstrating its capability to improve 3D mapping and pose estimation for automated robot delivery in ports.

\begin{table}[]
  \caption {Absolute Position Error (APE) of different methods in the large-scale port dataset [m]. The best results are in \textbf{BOLD}.}
  \label{tab:APE_port}
  \centering
  \scalebox{1.0}{
  \begin{tabular}{c|cccc}\hline
                    Sequence              & \makecell[c]{Initial\\ (SLAM + Loop)\citep{xu2022fast}} &PSS-GOSO    & \makecell[c]{Pose-graph\\~\citep{koide2019portable}} & \makecell[c]{HBA\\\citep{liu2023large}}\\\hline 
                    Port UGV & 8.63 & \textbf{1.31} & 4.33 & 3.31\\\hline 
  \end{tabular}
  }
\end{table}

\subsection{Evaluation on Proprietary Cross-platform Indoor-to-outdoor Dataset}

\begin{figure*}
  \centering
  \includegraphics[width=\linewidth]{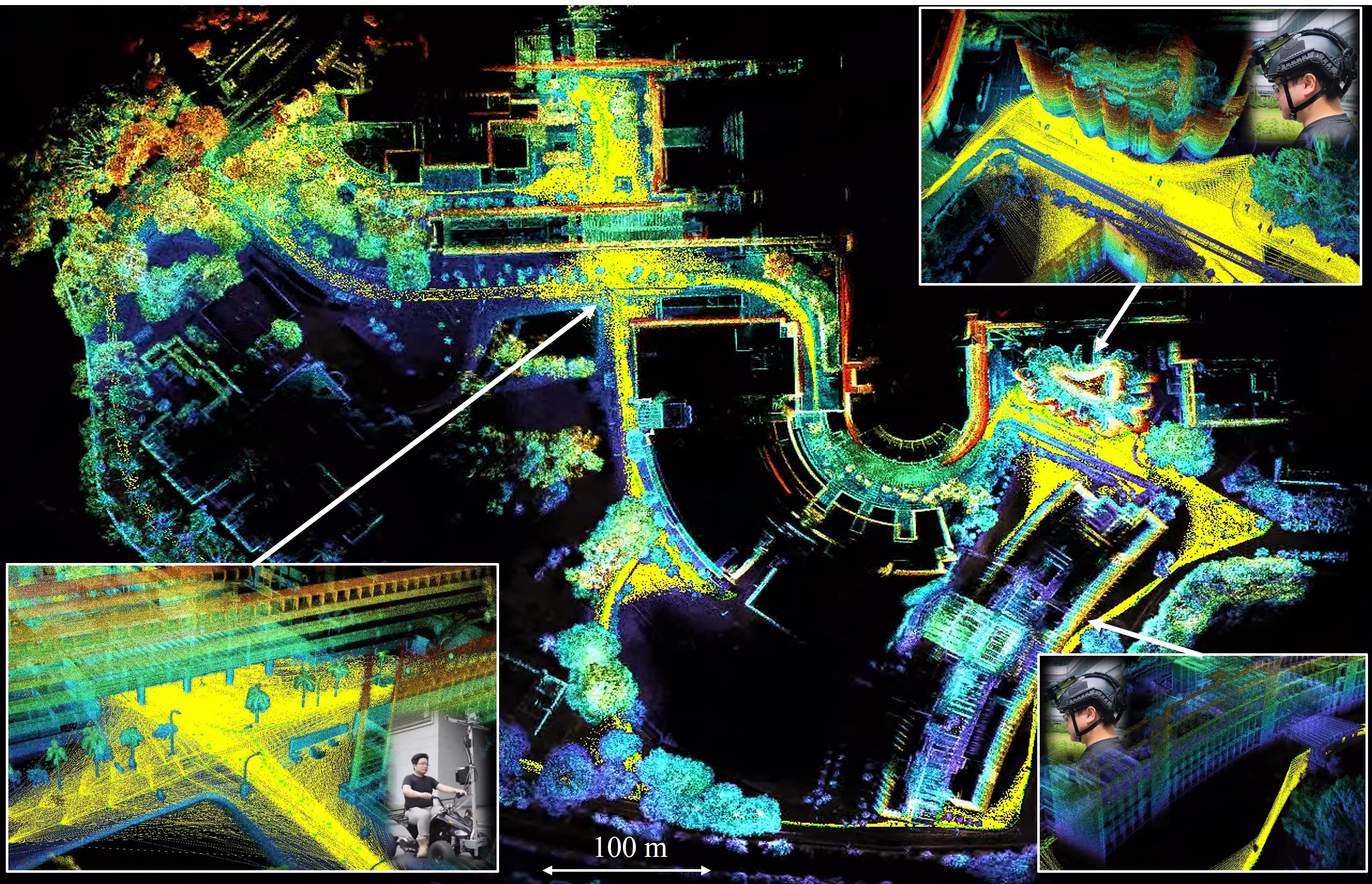}
  \caption{The point clouds and dense graphs in the proprietary indoor-to-outdoor dataset collected by multiple platforms. Edges of the graph are marked as yellow lines. The nodes of the graph are marked as green axes.}
  \label{fig:results_ntucampus}
\end{figure*}

\begin{figure}
  \centering
  \includegraphics[width=\linewidth]{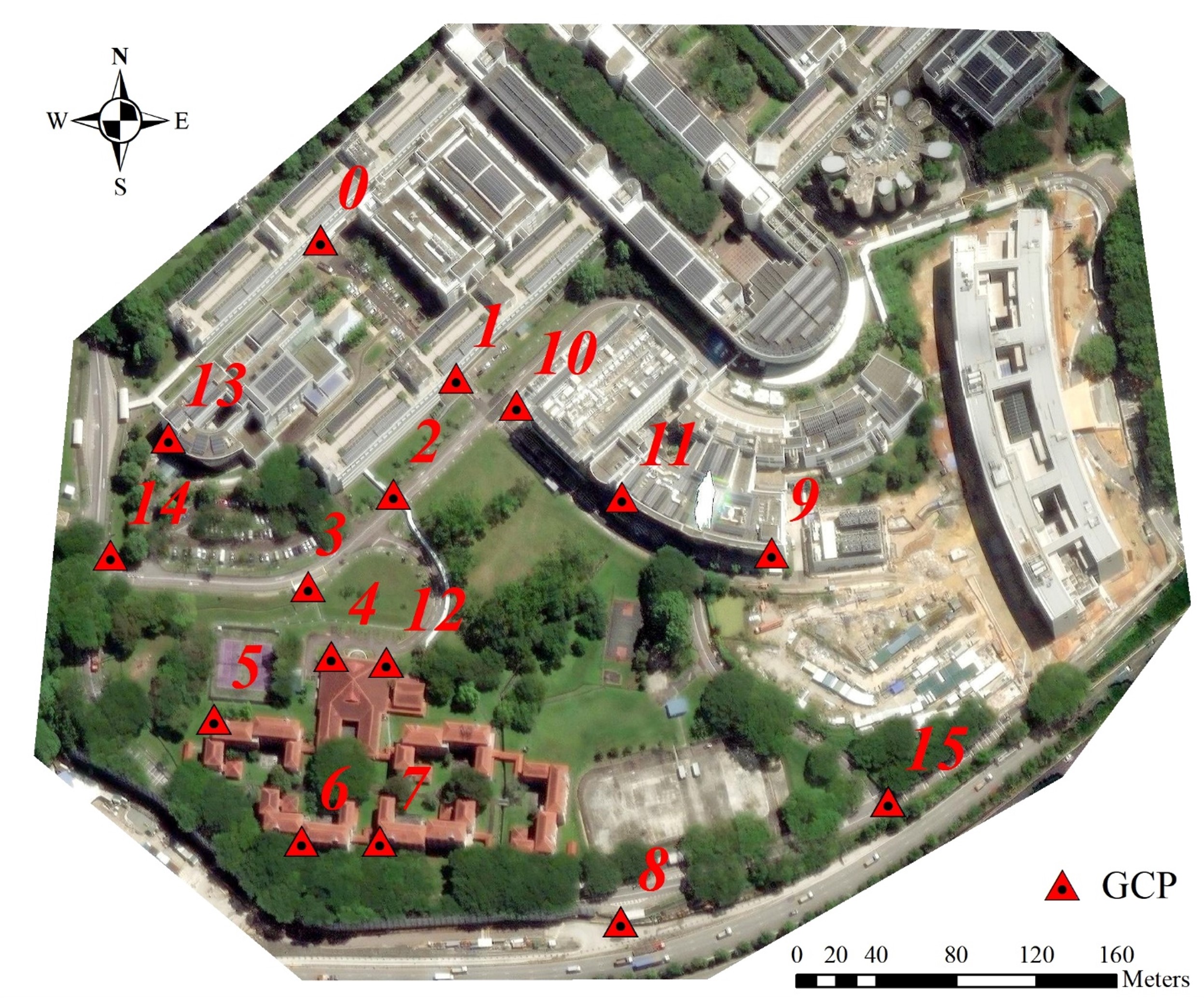}
  \caption{The ground control points surveyed using the total station.}
  \label{fig:gcps_ntucampus}
  \vspace{-15pt}
\end{figure}

\begin{figure}
  \centering
  \includegraphics[width=\linewidth]{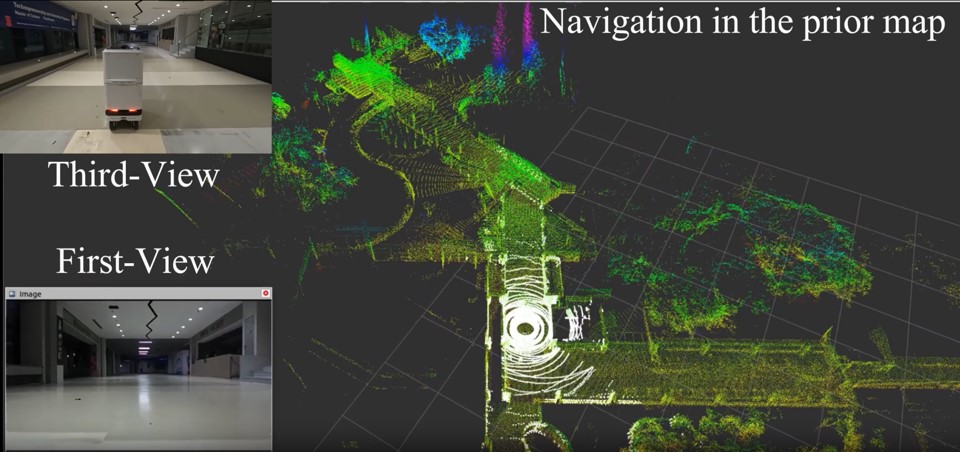}
  \caption{Navigation of the robot in the large-scale indoor-to-outdoor dataset. Demonstration video can be found on the project page.}
  \label{fig:navigation_demo}
\end{figure}

The 3D prior map is essential for applications such as outdoor-to-indoor surveillance and elderly health care, where accurate spatial information is crucial. Large-scale mapping tasks often involve cross-platform and multi-session data fusion, posing significant challenges for existing LBA methods. Traditional sequential splitting strategies, such as those used in HBA \citep{liu2023large}, are insufficient for addressing the complexities of multi-session and cross-platform data integration. In contrast, PSS-GOSO effectively tackles these challenges by partitioning the graph based on spatial relationships, allowing it to manage cross-platform and multi-session fusion tasks more efficiently. For our evaluation, initial pose estimates are derived using Fast-LIO and manual registration of submaps from different sessions. The dense graph and point clouds generated by PSS-GOSO are shown in Fig. \ref{fig:results_ntucampus}. To assess the accuracy of the point clouds produced by various methods, ground control points (GCPs) are collected using Leica RTC360 survey scanners as shown in Fig. \ref{fig:gcps_ntucampus}. The average errors of the GCPs, detailed in Table \ref{tab:ERROR_gcp}, indicate that PSS-GOSO reduced initial errors from 3.02 meters to 0.98 meters, demonstrating superior accuracy compared to existing pose-graph-based methods. This highlights PSS-GOSO’s effectiveness in generating high-accuracy point clouds for complex mapping scenarios involving diverse data sources and sessions. Moreover, the large-scale indoor-to-outdoor is used for the robot navigation as shown in Fig. \ref{fig:navigation_demo}. The real-time scanning data is matched with the prior map to get the accurate position for the robot, so the robot can navigate in large-scale complex environments for delivery and surveillance tasks. The demonstration video can be found on the project page.

\begin{table}[]
  \caption {GCP Errors of different methods in the multiple platforms dataset [m]. The best results are in \textbf{BOLD}.}
  \label{tab:ERROR_gcp}
  \centering
  \scalebox{0.9}{
  \begin{tabular}{c|cccc}\hline
                    Sequence              & \makecell[c]{Initial (SLAM + Regis)\\\citep{xu2022fast} } &PSS-GOSO    & \makecell[c]{Pose-graph\\~\citep{koide2019portable}} & \makecell[c]{HBA\\\citep{liu2023large}}\\\hline 
                    cross-platform & 3.02 & \textbf{0.98} & 1.76  & x \\\hline 
  \end{tabular}
  }
  \vspace{-15pt}
\end{table}

\subsection{Time efficiency of PSS-GOSO}

To evaluate the efficiency of PSS-GOSO and the impact of graph sparsification, we conduct an ablation study using a proprietary cross-platform dataset that includes both indoor and outdoor environments, which are listed in Table \ref{tab:time_efficiency}. Initially, we disabled the optimality-aware graph sparsification module within the GOSO framework, using the complete graph with all relationships included. This resulted in a raw graph with 65,926 edges and a processing time exceeding 4,800 seconds, highlighting the substantial computational burden. In contrast, when we enable the graph sparsification module in PSS-GOSO, the number of edges is reduced to 13,185 based on optimality criteria, significantly decreasing the processing time to 1,695 seconds. Despite the reduction in computational complexity, the accuracy of the final results remains high at 0.98 meters, comparable to the accuracy obtained with the fully connected graph. We also compare the processing time with the pose-graph-based method~\citep{koide2019portable}, which takes 1,218 seconds to get the final results. This comparison demonstrates that PSS-GOSO, through its graph sparsification module, effectively enhances processing efficiency while maintaining high accuracy, thus validating its effectiveness for large-scale LiDAR bundle adjustments and high-performance 3D mapping applications.

\begin{table}[]
  \caption {Time efficiency analysis of the PSS-GOSO on the Cross-platform indoor-to-outdoor dataset.}
  \label{tab:time_efficiency}
  \centering
  \scalebox{0.7}{
  \begin{tabular}{c|ccc}\hline
Methods                          & Number of Edges & Processing Time~(s) & Error of GCPs~(m)   \\\hline
Pose-graph~\citep{koide2019portable} & 65926 &  1218 & 1.76 \\
PSS-GOSO (w/o graph sparsifying) &    65926      & 4836 &        0.92     \\
PSS-GOSO                         &    13185        & 1695  &     0.98   \\\hline
\end{tabular}
  }
\end{table}

\section{Conclusion}\label{section:conclusion}

In this paper, we propose PSS-GOSO, a \textit{robust}, \textit{efficient}, and \textit{scalable} LiDAR Bundle Adjustment (LBA) method designed to enhance the accuracy of large-scale point clouds. To improve robustness, PSS-GOSO utilizes progressive spatial smoothing to extract reliable LiDAR correspondences for optimization. For increased efficiency, PSS-GOSO analyzes the graph structure of the raw bundle adjustment problem and selects the most critical edges to conserve computational resources. To address large-scale bundle adjustments, PSS-GOSO employs graph marginalization to decompose the problem into manageable subproblems. The performance of PSS-GOSO was evaluated across various platforms (handheld, legged, UGV, UAV) and diverse scenes, demonstrating its potential for high-accuracy 3D mapping. Optimized point clouds are successfully used as the prior map for robot last-mile delivery. In the near future, we plan to integrate the image data into the adjustment system to further enhance accuracy.

\section*{Appendix A: Auxiliary optimization}
\label{appendix_auxiliary}

To minimize Eq. \eqref{eq:normal}, we use the auxiliary function $\bm{\Xi}$ introduced by \cite{xu2011image}:  

\begin{equation}
    \begin{split}
    \underset{\hat{\mathbf{n}},|\hat{\mathbf{n}}|=1,\bm{\Xi}}{\mathrm{argmin}}\ \mathbf{G}(\hat{\mathbf{n}}_i) &= 1-\hat{\mathbf{n}}^{\top}_i\breve{\mathbf{n}}_i + \beta |\mathbf{D}(\hat{\mathbf{n}}_i)-\bm{\Xi}|^2 + \mu |\bm{\Xi}|_{0},
    \end{split}
\end{equation}
where the $\beta$ controls how quikly the auxiliary function $\bm{\Xi}$ approaches Eq. \eqref{eq:normal}, and is set to 0.01 at the begnining. With a initial guess of $\hat{\mathbf{n}}$, the above equation is optimized by fixing $\bm{\Xi}$ as follow:

\begin{equation}
    \begin{split}
    \underset{\bm{\Xi}}{\mathrm{argmin}}\ \mathbf{G}(\hat{\mathbf{n}}_i) &= \beta |\mathbf{D}(\hat{\mathbf{n}}_i)-\bm{\Xi}|^2 + \mu |\bm{\Xi}|_{0},
    \end{split} \label{eq_fix}
\end{equation}

The solution of Eq. \eqref{eq_fix} is given by:

\begin{equation}
    \left\{
    \begin{aligned}
        &\bm{\Xi}_j = 0, \frac{\mu}{\beta} > |\mathbf{D}(\hat{\mathbf{n}}_i)_j|^2\\
        &\bm{\Xi}_j = \mathbf{D}(\hat{\mathbf{n}}_i)_j, otherwise.
    \end{aligned} 
    \right.
\end{equation}

Next, $\bm{\Xi}$ is fixed to optimize the normal $\hat{\mathbf{n}}$ as follow:

\begin{equation}
    \begin{split}
    \underset{\hat{\mathbf{n}},|\hat{\mathbf{n}}|=1}{\mathrm{argmin}}\ \mathbf{G}(\hat{\mathbf{n}}_i) &= 1-\hat{\mathbf{n}}^{\top}_i\breve{\mathbf{n}}_i + \beta |\mathbf{D}(\hat{\mathbf{n}}_i)-\bm{\Xi}|^2.
    \end{split} \label{eq_quadratic}
\end{equation}

Eq. \ref{eq_quadratic} is quadratic and can be optimized using least square estimation. It should be noted that as $|\hat{\mathbf{n}}|=1$, $\hat{\mathbf{n}}$ is updated using Eq. \eqref{eq:normal_error}. The optimization of Eq. \eqref{eq_fix} and Eq. \eqref{eq_quadratic} iteratively with $\beta \leftarrow 2 \beta$ to make $\bm{\Xi}$ approach $\mathbf{D}(\hat{\mathbf{n}}_i)$.  

\section*{Appendix B: Relative Pose Residual and Covariance from Pair-wise Point Cloud Registration}
\label{appendix_covariance_reigstration}

For the edge $\mathbf{E}_k$, its first and second elements are the $\mathbf{E}_{k}^{0}$-th pose and $\mathbf{E}_{k}^1$-th pose, respectively. The relative pose residual ($\boldsymbol{\epsilon}_k$) and covariance ($\mathbf{\Omega}_k$) between the $\mathbf{E}_{k}^{0}$-th pose and $\mathbf{E}_{k}^1$-th pose are approximated using the pair-wise registration formulation. The $\mathbf{E}_{k}^{0}$-th pose and $\mathbf{E}_{k}^{1}$-th pose are denoted as $\mathbf{R}_{L_0},\mathbf{t}_{L_0}$ and $\mathbf{R}_{L_1},\mathbf{t}_{L_1}$ for convinience. The relative pose is denoted as $\mathbf{R}^{L_0}_{L_1}=\mathbf{R}^{\top}_{L_0}\mathbf{R}_{L_1},\mathbf{t}^{L_0}_{L_1}=\mathbf{R}^{\top}_{L_0}(-\mathbf{t}_{L_0}+\mathbf{t}_{L_1})$. Utilizing nearest searching and taking the point-to-point correspondeces $\{(\mathbf{P}^{L_0}_u,\mathbf{P}^{L_1}_u)\}^{U}_{u=1}$ as input, the registration function $\boldsymbol{\epsilon}^{\mathrm{reg}}_k$ for $\mathbf{E}_k$ is as follow. 
\begin{equation}
  \begin{split}
  \boldsymbol{\epsilon}^{\mathrm{reg}}_k=\sum_{u=1}^{U}(\mathbf{R}^{L_0}_{L_1}\mathbf{P}^{L_1}_u+\mathbf{t}^{L_0}_{L_1}-\mathbf{P}^{L_0}_u),
  \end{split}
\end{equation}
The Jacobian of $\boldsymbol{\epsilon}^{\mathrm{reg}}_k$ respect to the relative pose $\mathbf{R}^{L_0}_{L_1},\mathbf{t}^{L_0}_{L_1}$ is calculated by \eqref{eq:jacobian_regis}.

\begin{equation}
  \begin{split}
  \mathbf{J}^{\mathrm{reg}}_k=\sum_{u=1}^{U}\left[ \begin{matrix}
    -\left[ \mathbf{P}_{u}^{L_0} \right] _{\times}&		\mathbf{0}\\
    \mathbf{0}&		\mathbf{I}
  \end{matrix} \right]. \label{eq:jacobian_regis}
  \end{split}
\end{equation}
The covariance $\mathbf{\Omega}_k$ of the registration function $\boldsymbol{\epsilon}^{\mathrm{reg}}_k$ is calculated as follows.
\begin{equation}
  \begin{split}
  \mathbf{\Omega}_k=\mathbf{J}^{\mathrm{reg}\top}_k\mathbf{J}^{\mathrm{reg}}_k.
  \end{split}
\end{equation}
Without solving the registration function $\boldsymbol{\epsilon}^{\mathrm{reg}}_k$, the relative pose residual $\boldsymbol{\epsilon}_k$ with covariance $\mathbf{\Omega}_k$ for edge $\mathbf{E}_k$ could be derived as follow:
\begin{equation}
  \begin{split}
  \boldsymbol{\epsilon}_k=\left[ \begin{array}{c}
    (\hat{\mathbf{R}}^{L_0}_{L_1})^\top \mathbf{R}^\top_{L_0} \mathbf{R}_{L_1}\\
    -\hat{\mathbf{t}}^{L_0}_{L_1}+\mathbf{R}^\top_{L_0}(-\mathbf{t}_{L_0}+\mathbf{t}_{L_1})\\
  \end{array} \right] 
  \end{split}
\end{equation}

\section*{Appendix C: Jacobian of Progressive Spatial Smoothing (PSS) Factor}
\label{appendix_pss}
The Jacobian of $\sigma _{i,j}$ in \eqref{eq:pss_residual} with respect to the $i^{th}$ pose can be derived using the chain rule as follows:
\begin{equation}
\begin{split}
    &\frac{\partial \sigma _{i,j}}{\partial \hat{\mathbf{t}}_i}=\frac{\partial \sigma _{i,j}}{\partial \hat{\mathbf{p}}_{j}^{S}}\frac{\partial \hat{\mathbf{p}}_{j}^{S}}{\partial \hat{\mathbf{p}}_{i}^{W}}\frac{\partial \hat{\mathbf{p}}_{i}^{W}}{\partial \hat{\mathbf{t}}_i}=-\frac{\partial \sigma _{i,j}}{\partial \hat{\mathbf{p}}_{i}^{S}}\hat{\mathbf{M}}_i,
    \\
    &\frac{\partial \sigma _{i,j}}{\partial \hat{\bm{\theta}}_i}=\frac{\partial \sigma _{i,j}}{\partial \hat{\mathbf{p}}_{j}^{S}}\frac{\partial \hat{\mathbf{p}}_{j}^{S}}{\partial \hat{\mathbf{p}}_{i}^{W}}\frac{\partial \hat{\mathbf{p}}_{i}^{W}}{\partial \hat{\bm{\theta}}_i}=\frac{\partial \sigma _{i,j}}{\partial \hat{\mathbf{p}}_{i}^{S}}\hat{\mathbf{M}}_i\left[ \hat{\mathbf{R}}_i\mathbf{p}_{i}^{L} \right] _{\times},\\
    &\frac{\partial \sigma _{i,j}}{\partial \hat{\mathbf{p}}_{j}^{S}}=\left[ 2\mathbf{\alpha }_{0}^{i}\hat{x}_{j}^{S}+\mathbf{\alpha }_{2}^{i}\hat{y}_{j}^{S}+\mathbf{\alpha }_{3}^{i},2\mathbf{\alpha }_{1}^{i}\hat{y}_{j}^{S}+\mathbf{\alpha }_{2}^{i}\hat{x}_{j}^{S}+\mathbf{\alpha }_{4}^{i},-1 \right].     
\end{split}
\end{equation}
Similarly, the Jacobian of $\sigma _{i,j}$ respect with the $j^{th}$ pose is as follows: 
\begin{equation}
    \begin{split}
    &\frac{\partial \sigma _{i,j}}{\partial \hat{\mathbf{t}}_j}=\frac{\partial \sigma _{i,j}}{\partial \hat{\mathbf{p}}_{j}^{S}}\frac{\partial \hat{\mathbf{p}}_{j}^{S}}{\partial \hat{\mathbf{p}}_{j}^{W}}\frac{\partial \hat{\mathbf{p}}_{j}^{W}}{\partial \hat{\mathbf{t}}_j}=\frac{\partial \sigma _{i,j}}{\partial \hat{\mathbf{p}}_{j}^{S}}\hat{\mathbf{M}}_i,
    \\
    &\frac{\partial \sigma _{i,j}}{\partial \hat{\bm{\theta}}_j}=\frac{\partial \sigma _{i,j}}{\partial \hat{\mathbf{p}}_{j}^{S}}\frac{\partial \hat{\mathbf{p}}_{j}^{S}}{\partial \hat{\mathbf{p}}_{j}^{W}}\frac{\partial \hat{\mathbf{p}}_{j}^{W}}{\partial \hat{\bm{\theta}}_j}=-\frac{\partial \sigma _{i,j}}{\partial \hat{\mathbf{p}}_{j}^{S}}\hat{\mathbf{M}}_i\left[ \hat{\mathbf{R}}_j\mathbf{p}_{j}^{L} \right] _{\times}.       
    \end{split}
\end{equation}

\setlength{\bibsep}{2pt} 
\renewcommand{\bibfont}{\small}
\bibliography{cas-refs} 
\bibliographystyle{IEEEtranN}

%\vfill

\end{document}